\documentclass[letterpaper]{article} 
\usepackage{aaai24}  
\usepackage{times}  
\usepackage{helvet}  
\usepackage{courier}  
\usepackage[hyphens]{url}  
\usepackage{graphicx} 
\usepackage{natbib}  
\usepackage{caption} 
\usepackage{algorithm}
\usepackage{algorithmic}
\usepackage{amsmath}
\usepackage{amsfonts,amssymb} 

\usepackage{newfloat}
\usepackage{listings}
\DeclareCaptionStyle{ruled}{labelfont=normalfont,labelsep=colon,strut=off} 
\floatstyle{ruled}
\newfloat{listing}{tb}{lst}{}
\floatname{listing}{Listing}
\usepackage{soul}
\usepackage{xcolor}

\newcommand{\shorttitle}[0]{{Sparse3D}}

\title{\shorttitle{}: Distilling Multiview-Consistent Diffusion for Object Reconstruction from Sparse Views}
\author{
    Zi-Xin Zou\textsuperscript{\rm 1}, Weihao Cheng\textsuperscript{\rm 2}, Yan-Pei Cao\textsuperscript{\rm 2}, Shi-Sheng Huang\textsuperscript{\rm 3}, \\ Ying Shan\textsuperscript{\rm 2}, Song-Hai Zhang\textsuperscript{\rm 1\dag}
}
\affiliations{
    \textsuperscript{\rm 1}BNRist, Tsinghua University\;
    \textsuperscript{\rm 2}ARC Lab, Tencent PCG \;
    \textsuperscript{\rm 3}Beijing Normal University \\
}

\usepackage{bibentry}
\usepackage{multicol}
\usepackage{lipsum}

\begin{document}

\maketitle

\begin{abstract}
Reconstructing 3D objects from extremely sparse views is a long-standing and challenging problem. 
While recent techniques employ image diffusion models for generating plausible images at novel viewpoints or for distilling pre-trained diffusion priors into 3D representations using score distillation sampling (SDS), these methods often struggle to simultaneously achieve high-quality, consistent, and detailed results for both novel-view synthesis (NVS) and geometry.
In this work, we present \emph{\shorttitle{}}, a novel 3D reconstruction method tailored for sparse view inputs. Our approach distills robust priors from a multiview-consistent diffusion model to refine a neural radiance field.
Specifically, we employ a controller that harnesses epipolar features from input views, guiding a pre-trained diffusion model, such as Stable Diffusion, to produce novel-view images that maintain 3D consistency with the input.
By tapping into 2D priors from powerful image diffusion models, our integrated model consistently delivers high-quality results, even when faced with open-world objects.
To address the blurriness introduced by conventional SDS, we introduce category-score distillation sampling (C-SDS) to enhance detail.
We conduct experiments on CO3DV2 which is a multi-view dataset of real-world objects.
Both quantitative and qualitative evaluations demonstrate that our approach outperforms previous state-of-the-art works on the metrics regarding NVS and geometry reconstruction.

\end{abstract}

\section{Introduction}

\begin{figure}[t]
    \centering
    \includegraphics[scale=0.5]{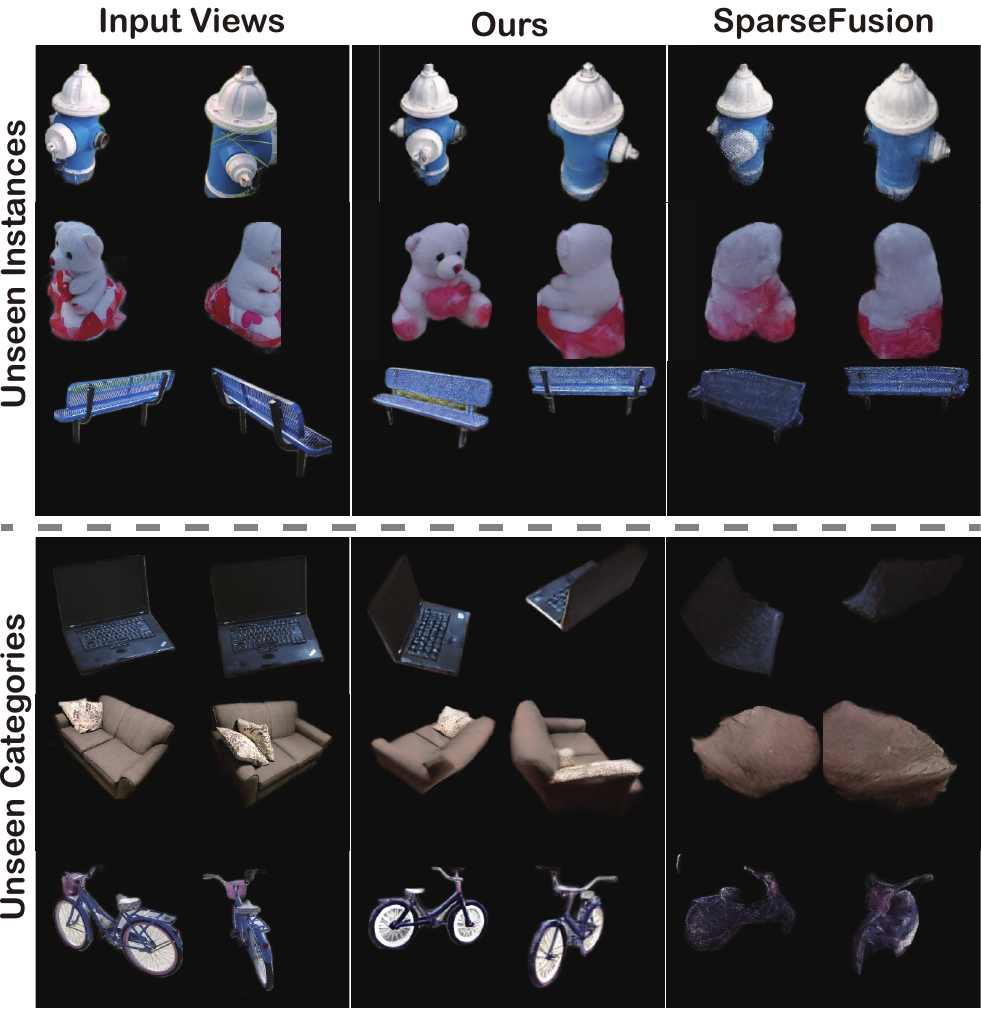}
    \caption{\textbf{Novel-view synthesis from two input views using our \shorttitle{} and SparseFusion.} Our approach can achieve higher-quality images with more details for unseen instances, especially for the unobserved regions of them (e.g., the left face of the teddybear). Furthermore, our approach can generalize to some unseen categories without any further finetuning, while SparseFusion fails.}
    \label{fig:teaser}
\end{figure}


Reconstructing 3D objects from sparse-view images remains a pivotal challenge in the realms of computer graphics and computer vision. This technique has a wide range of applications such as Augmented and Virtual Reality (AR/VR). The advent of the Neural Radiance Field (NeRF) and its subsequent variants has catalyzed significant strides in geometry reconstruction and novel-view synthesis, as delineated in recent studies~\cite{DBLP:conf/eccv/MildenhallSTBRN20,DBLP:conf/nips/WangLLTKW21,DBLP:conf/nips/YarivGKL21}. However, NeRFs exhibit limitations when operating on extremely sparse views, specifically with as few as 2 or 3 images. In these scenarios, the synthesized novel views often suffer in quality due to the limited input observations.

Existing methods for sparse-view reconstruction typically leverage a generalizable NeRF model, pre-trained on multi-view datasets, to infer 3D representations from projected image features~\cite{DBLP:conf/cvpr/YuYTK21,SRF}. 
However, these approaches tend to regress to the mean, failing to produce perceptually sharp outputs, especially in intricate details.
\let\thefootnote\relax\footnote{\hspace*{-1.8em}$^\dag$ corresponding author}
To produce plausible results, either in terms of geometry or appearance, from limited observations, several studies have turned to image generation models, such as the diffusion model~\cite{DBLP:conf/cvpr/RombachBLEO22}, to ``imagine'' unseen views based on provided images~\cite{DBLP:journals/corr/abs-2304-02602,zhou2023sparsefusion}.
For example, Zero123~\cite{liu2023zero1to3} trains a view-conditioned diffusion model on a large synthetic dataset and achieves impressive results. However, their generated images across different views may not be consistent.
Thus, while these view-conditioned diffusion models can produce satisfactory images, their quality and generalization ability are often constrained by the scarcity of posed image datasets.
Large-scale image diffusion models~\cite{DBLP:conf/icml/RameshPGGVRCS21,DBLP:conf/nips/SahariaCSLWDGLA22,DBLP:conf/cvpr/RombachBLEO22}, which are pre-trained on billions of 2D images~\cite{DBLP:conf/nips/SchuhmannBVGWCC22}, excel in generating high-quality and diverse images. However, despite the diverse, general capability of such models, in 3D reconstruction tasks, users need to synthesize specific instances that are coherent with user-provided input images. Even with recent model customization methods~\cite{kumari2023multi,ruiz2023dreambooth,gal2022image}, they prove unwieldy and often fail to produce the specific concept with sufficient fidelity.
Consequently, the potential of merging the capabilities of pre-trained large image diffusion models with the viewpoint and appearance perception of specific instances remains an open avenue of exploration.

In contrast to directly generating images at novel views, some recent works explore distilling the priors of pre-trained diffusion models into a NeRF (neural radiance field) framework. This approach facilitates 3D-consistent novel-view synthesis and allows for mesh extraction from the NeRF.
Notable works such as DreamFusion~\cite{DBLP:conf/iclr/PooleJBM23} and SJC~\cite{sjc} employ score distillation sampling (SDS) to harness off-the-shelf diffusion models for text-to-3D generation. However, a persistent challenge with SDS is the production of blurry and oversaturated outputs, attributed to noisy gradients, which in turn compromises the quality of NeRF reconstructions.

In this work, we present \emph{\shorttitle{}}, a novel 3D reconstruction approach designed to reconstruct high-fidelity 3D objects from sparse and posed input views. Our method hinges on two pivotal components: \textbf{(1)} a diffusion model that ensures both multiview consistency and fidelity to user-provided input images while retaining the powerful generalization capabilities of Stable Diffusion~\cite{DBLP:conf/cvpr/RombachBLEO22}, and \textbf{(2)} a category-score distillation sampling (C-SDS) strategy.
At its core, we distill the priors from our fidelity-preserving, multiview-consistent diffusion model into the NeRF reconstruction using an enhanced category-score distillation sampling.
Specifically, for the multiview-consistent diffusion model, we propose to utilize an epipolar controller to guide the off-the-shelf Stable Diffusion model to generate novel-view images that are 3D consistent with the content of input images. 
Notably, by fully harnessing the 2D priors present in Stable Diffusion, our model exhibits robust generalization capabilities, producing high-quality images even when confronted with open-world, unseen objects.
To overcome the problem of blurry, oversaturated, and non-detailed results caused by SDS during NeRF reconstruction, we draw inspiration from VSD~\cite{DBLP:journals/corr/abs-2305-16213} and propose a category-score distillation sampling strategy (C-SDS). Additionally, two perception losses between the one-step estimation image from our diffusion model and the rendering image are employed for better results, without incurring much extra computational cost.

We evaluate \emph{Sparse3D} on the Common Object in 3D (CO3DV2) dataset and benchmark it against existing approaches.
The results show that our approach outperforms state-of-the-art techniques in terms of the quality of both synthesized novel views and reconstructed geometry.
Importantly, \emph{Sparse3D} exhibits superior generalization capabilities, particularly for object categories not present in the training domain.

\section{Related Works}
\subsection{Multi-view 3D Reconstruction}
Multi-view 3D reconstruction is a long-standing problem with impressive works such as traditional Structure-from-Motion (S\emph{f}M)~\cite{DBLP:conf/cvpr/SchonbergerF16} or Multi-view-Stereo (MVS)~\cite{DBLP:conf/eccv/SchonbergerZFP16}, and recent learning based approaches~\cite{DBLP:conf/eccv/YaoLLFQ18,DBLP:conf/cvpr/YuG20}.
The success of NeRF~\cite{DBLP:conf/eccv/MildenhallSTBRN20,DBLP:journals/tog/MullerESK22} has led to impressive outcomes in novel-view synthesis and geometric reconstruction.
However, these methods still struggle to produce satisfactory results for extremely sparse view scenarios. Subsequent works proposed to use regularization (semantic~\cite{DBLP:conf/iccv/JainTA21}, frequency~\cite{Yang2023FreeNeRF}, geometry and appearance~\cite{DBLP:conf/cvpr/NiemeyerBMS0R22}) and geometric priors (e.g. depth~\cite{DBLP:conf/cvpr/DengLZR22,DBLP:conf/cvpr/RoessleBMSN22} or normal~\cite{DBLP:conf/nips/YuPNS022}) but remain to be inadequate for view generation in unobserved regions, due to the essential lack of scene priors. 

\subsection{Generalizable Novel-view Synthesis}
For generalizable novel-view synthesis using NeRF, some approaches utilize projected features of the sampling points in volumetric rendering~\cite{DBLP:conf/cvpr/YuYTK21,DBLP:conf/cvpr/WangWGSZBMSF21,SRF}, or new neural scene representations, such as Light Field Network~\cite{DBLP:conf/cvpr/SuhailESM22,DBLP:conf/eccv/SuhailESM22} or Scene Representation Transformer~\cite{DBLP:conf/cvpr/SajjadiMPBGRVLD22}  for better generalizable novel-view synthesis.
Subsequent researches~\cite{DBLP:conf/eccv/KulhanekDSB22,DBLP:journals/corr/abs-2304-02602,DBLP:journals/corr/abs-2306-03414} propose to further utilize generative models (e.g. VQ-VAE~\cite{DBLP:conf/nips/OordVK17} and diffusion model~\cite{DBLP:conf/cvpr/RombachBLEO22}) to generate unseen images.
However, these methods didn't have any 3D-aware scene priors, which limits their potential applications.
In this paper, we leverage the feature map from a generalizable renderer to guide a pre-trained diffusion model to generate multiview-consistent images, and then distill the diffusion prior into NeRF reconstruction for both novel-view synthesis and geometry reconstruction.

\begin{figure*}[t]
    \centering
    \includegraphics[width=\linewidth]{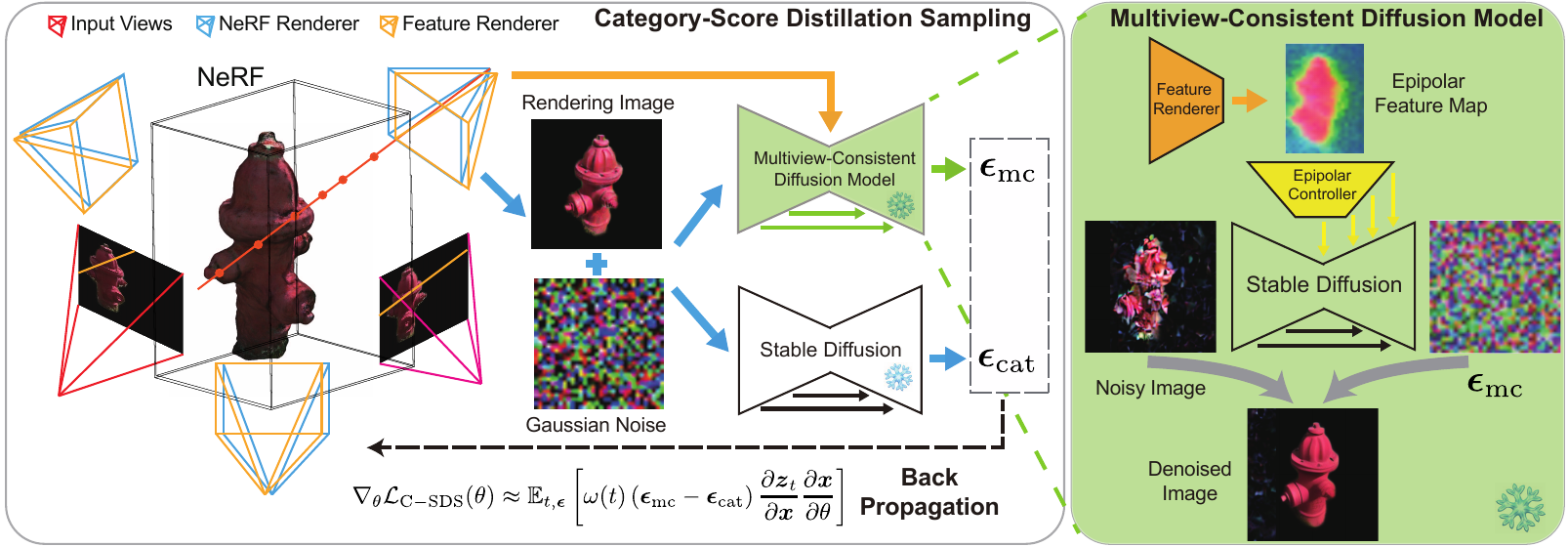}
    \caption{\textbf{Overview of \shorttitle{}}. Our approach consists of two key components: a multiview-consistent diffusion model and a category-score distillation sampling. We utilize epipolar feature map to control the Stable Diffusion model to generate images consistent with the content of input images, serving as a multiview-consistent diffusion model. Based on such a model, we propose a category-score distillation sampling (C-SDS) strategy to achieve more detailed results during NeRF reconstruction. 
    }
    \label{fig:pipeline}
\end{figure*}

\subsection{3D Generation with 2D Diffusion Model}
Diffusion denoising probabilistic models have brought a boom of generation tasks for 2D images and 3D contents in recent years.
Inspired by early works which use CLIP embedding~\cite{DBLP:conf/iccv/JainTA21,DBLP:conf/cvpr/WangCH0022,DBLP:conf/cvpr/JainMBAP22} or GAN~\cite{DBLP:conf/iclr/PanDLLL21} to regularize the NeRF, DreamFusion~\cite{DBLP:conf/iclr/PooleJBM23} and SJC~\cite{sjc} propose a score distillation sampling (SDS) strategy to guide the NeRF optimization for impressive text-to-3D generation. ProlificDreamer~\cite{DBLP:journals/corr/abs-2305-16213} proposes variational score distillation (VSD) for more high-fidelity and diverse text-to-3D generation.
Magic3D~\cite{lin2023magic3d} improves the 3D generation quality by a two-stage coarse-to-fine strategy.
To generate 3D results consistent with the input image observation, subsequent works leverage textual-inversion~\cite{melaskyriazi2023realfusion} or denoised-CLIP loss with depth prior~\cite{tang2023make}.
When additional geometry prior are available (e.g. point clouds from Point-E~\cite{DBLP:journals/corr/abs-2212-08751}), some works~\cite{DBLP:journals/corr/abs-2303-07937,DBLP:journals/corr/abs-2307-13908} can produce more 3D consistent creation.
In addition to lifting a pre-trained diffusion model, Zero123~\cite{liu2023zero1to3}, SparseFusion~\cite{zhou2023sparsefusion} and NerfDiff~\cite{DBLP:journals/corr/abs-2302-10109} train a viewpoint-conditioned diffusion model and achieve impressive results.
Instead of training a diffusion model or directly lifting a pre-trained diffusion model, our approach leverages both the advantages of them to train a multiview-consistent diffusion model, with a category-score distillation sampling to improve the results of SDS for more details.

\section{Method}
Given $N$ input images $\{I_n\}_{n=1}^{N}$ of an object with corresponding camera poses $\{T_n\}_{n=1}^{N}$, where $N$ can be as few as 2, our goal is to reconstruct a neural radiance field (NeRF), enabling generalizable novel view synthesis and high-quality surface reconstruction.
To realize this goal, we propose \shorttitle{}, which 
distills a multiview-consistent diffusion model prior into the NeRF representation of an object, using a category-score distillation sampling (C-SDS) strategy.
Figure~\ref{fig:pipeline} shows the overview of our approach.
The multiview-consistent diffusion model extracts epipolar features from sparse input views and uses a control network to guide the Stable Diffusion model to generate novel-view images that are faithful to the object shown in the images. A NeRF is then reconstructed with the guidance of the diffusion model.
To overcome the blurry problem that occurred in SDS, we propose C-SDS.
Benefiting from C-SDS, the gradients conditioned on category prior maintain the optimization with a tightened region of the search space, leading to more detailed results.
Finally, our approach achieves more consistent and high-quality results of novel-view synthesis and geometry reconstruction. 
We introduce the details of the multiview-consistent diffusion model and the C-SDS-based NeRF reconstruction in the following subsections.

\subsection{Multiview-Consistent Diffusion Model} 
\begin{figure}
    \centering
    \includegraphics[scale=0.65]{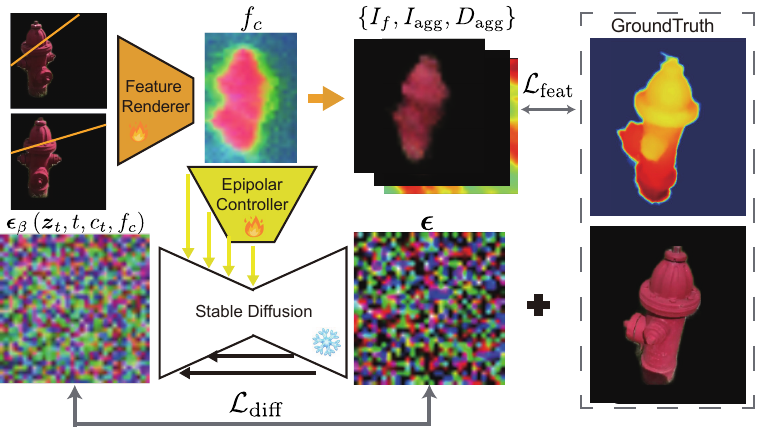}
    \caption{\textbf{Multiview-consistent diffusion model.} Our multiview-consistent diffusion model comprises a feature renderer, an epipolar controller, and a Stable Diffusion model.}
    \label{fig:train}
\end{figure}

Our diffusion model consists of a feature renderer, an epipolar controller, and a Stable Diffusion model, where the epipolar controller and the Stable Diffusion model together constitute the noise predictor $\boldsymbol{\epsilon}_{\beta}$, as shown in Figure~\ref{fig:train}.
The feature renderer $g_{\psi}$ takes a set of posed images and viewpoint $\pi$ as input, subsequently outputting an epipolar feature map $f_c=g_{\psi}(\pi, I_1,...,I_n,T_1,...,T_n)$, which serves as the input for the epipolar controller.
To unify the pre-trained diffusion model and multiview-consistent perception ability for a specific object, we draw inspiration from ControlNet~\cite{zhang2023adding}. 
ControlNet enables image generation controlled by conditional inputs (such as edge maps, segmentation maps, and depth maps).
Instead, we use the epipolar feature map to guide a pre-trained diffusion model to generate images consistent with the content of input images from various viewpoints.
To align the feature space of the feature renderer and controller, we use a convolution layer to map the features before feeding them into the controller.

\paragraph{Feature Renderer.}
Previous works acquire the feature map $f_c$ through rendering from Triplane~\cite{DBLP:journals/corr/abs-2302-10109}, 3D Volume~\cite{DBLP:journals/corr/abs-2304-02602} or epipolar feature transformer~\cite{zhou2023sparsefusion}.
In this paper, we adapt epipolar feature transformer (EFT) following ~\cite{zhou2023sparsefusion}.
The EFT, derived from GPNR~\cite{DBLP:conf/eccv/SuhailESM22}, learns a network $g_{\psi}$ to predict color of given ray $\boldsymbol{r}$ from input images. The rendering process primarily involves three transformers, which output attention weights used to blend colors over input views and epipolar lines for the final prediction. We recommend readers to refer to ~\cite{zhou2023sparsefusion} and ~\cite{DBLP:conf/eccv/SuhailESM22} for the details of epipolar feature rendering.
We implement two modifications to the EFT for improved results:
(1) a mask embedding and a relative camera transformation embedding are concatenated with other transformer token features. 
(2) To enhance generalizability and achieve better geometry awareness, we also obtain the aggregated color $I_{\text{agg}}$ and depth images $D_{\text{agg}}$ by attention weights of transformers to compute loss. 

\paragraph{Epipolar Controller.}
{Given} feature maps $f_c$ rendered at {arbitrary} viewpoints, we propose to {learn} an epipolar controller to guide a pre-trained diffusion model to generate multiview-consistent images with high quality.
Our epipolar controller takes epipolar feature map $f_c$ and category text prompt $c_t$ as input, subsequently outputting the latent features that are fused with the latent features of Stable Diffusion.
We also employ a convolution layer to align the dimensions of the feature map and epipolar controller input.
Rather than training a new diffusion model, we hope to retain the rich 2D priors from Stable Diffusion.
Consequently, we jointly train our epipolar controller and feature renderer, while keeping the parameters of Stable Diffusion fixed.
On the one hand, by utilizing the feature map, which contains implicit information about the appearance of the specific object and perception of the observation viewpoint, we can control a pre-trained text-to-image diffusion model to generate images consistent with the content of input images from different viewpoints.
On the other hand, our diffusion model inherits the high-quality image generation capabilities from Stable Diffusion, and the additional category prior in the text domain can also enhance the multiview consistency. 
Furthermore, these priors also enable our model to generalize to open-world unseen categories.

\paragraph{Training.}
Finally, we jointly train the feature renderer and the epipolar controller by the following objective function:
\begin{equation}
\mathcal{L}=\mathcal{L}_{\text{feat}}+\mathcal{L}_{\text{diff}}
\end{equation}
where $\mathcal{L}_{\text{feat}}$ is {the loss} for feature renderer {and} $\mathcal{L}_{\text{diff}}$ is {the loss} for epipolar controller.
While the feature map primarily serves as input for the controller in our pipeline, we also supervise it with color images and depth images to enhance its perception of appearance, observation viewpoints, and geometry awareness.
For a query ray $r$ from novel view when given input images, we decode the color $I_{f}$ from the feature map and supervise it using ground-truth color values. Additionally, to improve generalizability and geometry awareness, we employ a Mean Squared Error (MSE) loss on aggregated color $I_{\text{agg}}$ and depth $D_{\text{agg}}$. We then formulate the objective function as follows:
\begin{equation}
\begin{aligned}
    \mathcal{L}_{\text{feat}}=&\sum_{r}^{}||I_{f}(r)-I(r)||^2 + ||I_
    {\text{agg}}(r)-I(r)||^2 \\& +||D_{\text{agg}}(r)-D(r)||^2
\end{aligned}
\end{equation}
where $I(r)$ and $D(r)$ are 
ground-truth color and depth image respectively.

The diffusion model learns a conditional noise predictor to estimate the denoising score by adding Guassian-noise $\boldsymbol{\epsilon}$ to clean data in $T$ timesteps. 
We minimize the noise prediction error at randomly sampled timestep $t$. The objective of the diffusion model conditioned on text prompt $c_t$ (we use the category name as the conditioned text prompt, e.g. ``hydrant'') and feature map $f_c$ is given by:
\begin{equation}
    \mathcal{L}_{\text{diff}} = \mathbb{E}_{\boldsymbol{\epsilon}\sim\mathcal{N}(0,1)}||\boldsymbol{\epsilon}-\boldsymbol{\epsilon}_{\beta}(\boldsymbol{z}_t, t, c_t, f_c)||^2
\end{equation}
where $\epsilon_{\beta}$ is the conditional noise predictor of our diffusion model.

\subsection{NeRF Reconstruction with C-SDS}
Building on our multiview-consistent diffusion model, we aim to optimize a neural radiance field (NeRF) parameterized with $\theta$, from which more 3D-consistent novel-view synthesis and underlying explicit geometry can be derived.
Then to overcome the problem of blurry and non-detailed results in SDS, we propose a category-score distillation sampling (C-SDS) strategy.

\begin{figure*}[t]
    \centering
    \includegraphics[scale=0.5]{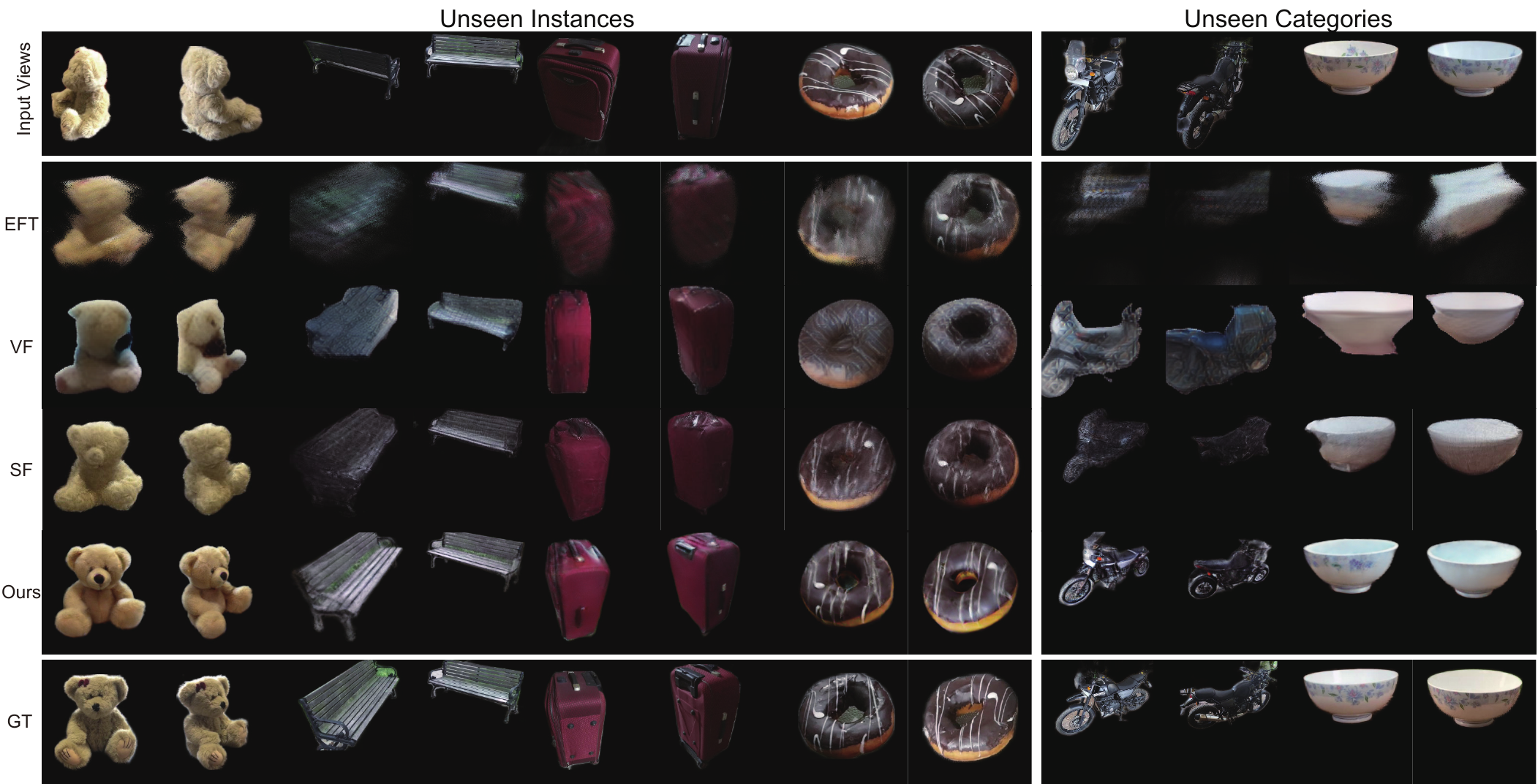}
    \caption{\textbf{Qualitative comparison of novel-view synthesis when given 2 input views.} Our approach achieves both high quality and more details of novel-view images compared to the others (e.g., the face of the teddybear), whenever with unseen instances and unseen categories.}
    \label{fig:2view}
\end{figure*}

\paragraph{Category-Score Distillation Sampling.} To overcome the problem of SDS, we draw inspiration from VSD~\cite{DBLP:journals/corr/abs-2305-16213} and propose a C-SDS for more detailed outcomes as follows:
\begin{equation}
\nabla_\theta \mathcal{L}_{\mathrm{C-SDS}}(\theta) \approx \mathbb{E}_{t, \boldsymbol{\epsilon}}\left[\omega(t)\left(\boldsymbol{\epsilon}_{\text{mc}}-\boldsymbol{\epsilon_{\text{cat}}}\right) \frac{\partial\boldsymbol{z}_t}{\partial \boldsymbol{x}} \frac{\partial \boldsymbol{x}}{\partial \theta}\right] \\
\end{equation}
where $\boldsymbol{\epsilon_{\text{mc}}}=\boldsymbol{\epsilon}_{\beta}(\boldsymbol{z}_t, t, c_t, f_c)$ is the predicted noise by our multiview-consistent diffusion model, $\boldsymbol{\epsilon_{\text{cat}}}=\boldsymbol{\epsilon_{\text{sd}}}\left(\boldsymbol{z}_t; t, c_t\right)$ is the predicted noise by Stable Diffusion conditioned text prompt of category $c_t$. 
And $\omega(t)$ is a weighting function that depends on the timestep $t$.

Instead of employing a Gaussian noise as SDS does, we replace it with an estimation $\boldsymbol{\epsilon_{\text{cat}}}$ incorporating category prior from Stable Diffusion.
By providing an approximation of the estimation of the score function of the distribution on rendering images with category prior, our C-SDS can deliver a better gradient with a tightened region of the search space, resulting in more detailed outputs.
SDS relies on high classifier-free guidance (CFG, i.e. 100) to achieve a better convergence, but such high CFG may lead to over-saturation and over-smooth problems~\cite{DBLP:conf/iclr/PooleJBM23}.
One reason for the reliance on large CFG is that the pre-trained text-to-image diffusion model may not obtain multiview-consistent image generation at novel views, thus providing a noisy estimation across different viewpoints.
In our experiment, when using a more multiview-consistent diffusion model, it can work with a small CFG (i.e. 7.5).
However, the results still suffer from blurry and non-detailed outputs, as the update gradient is not accurate enough.
ProlificDreamer utilizes a low-rank adaption (LoRA) of a pre-trained diffusion model to estimate the score function of the distribution on rendered images.
In our experiment, we find that it is hard for LoRA to provide good estimation during our instance-specific optimization.
Therefore, our proposed C-SDS offers a simple yet effective way to estimate the score function of the distribution on rendered images for more detailed results.

\paragraph{One-step Estimation from Diffusion Model.} 
The predicted noise from the diffusion model can be used not only in C-SDS but also to estimate its one-step denoising image without requiring much extra computation:
\begin{equation}
\begin{aligned}
&\boldsymbol{z}_{\text{1step}}=\frac{1}{\sqrt{\bar{\alpha}_t}}\left(\boldsymbol{z}_t-\sqrt{1-\bar{\alpha}_t} \boldsymbol{\epsilon}_\beta\left(\boldsymbol{z}_t, t, c_t, f_c\right)\right), \\
&\boldsymbol{x_{\text{1step}}}=\mathcal{D}(\boldsymbol{z_{\text{1step}}})
\end{aligned}
\end{equation}
where $\mathcal{D}$ is the decoder of Stable Diffusion which decodes latent features to image space. 
Although its one-step estimation may be blurry and sometimes inaccurate, making it unsuitable for performing pixel-wise loss, we can leverage it to provide an additional regularization term by using perceptual distance. We find that the perception regularization from one-step estimation improves the metrics of results. Specifically, we employ two perceptual losses, which include LPIPS loss~\cite{DBLP:conf/cvpr/ZhangIESW18} and contextual loss ~\cite{DBLP:conf/eccv/MechrezTZ18} to formulate the perception regularization from one-step estimation image:
\begin{equation}
\mathcal{L}_{\text{perp}}=\lambda_{p}\mathcal{L}_{\text{lpips}}(I,\boldsymbol{x}_{1step})+\lambda_{c}\mathcal{L}_{\text{contextual}}(I,\boldsymbol{x}_{1step})
\end{equation}

\paragraph{Reference Supervision.}
In addition to the guidance of diffusion priors at novel views, we use the reference input images $I$ with their masks $M$ to encourage the consistent appearance with the input images:
\begin{equation}
    \mathcal{L}_{\text{ref}}=\lambda_{r}||(\hat{I}-I)*\hat{M}||_2^2+\lambda_{m}||\hat{M}-M||_2^2
\end{equation}
where $\hat{I}$ and $\hat{M}$ are rendering image and mask, respectively.

\begin{table*}[t]
    \centering
    \small
    \begin{tabular}{c|cccccc|cccccc}
    \hline
        & \multicolumn{6}{c}{Unseen Instances - 2 views} & \multicolumn{6}{c}{Unseen Instance - 3 views} \\ \hline
        & PSNR $\uparrow$ & SSIM $\uparrow$ & LPIPS $\downarrow$ & FID $\downarrow$ & CLIP $\uparrow$ & DISTS $\downarrow$ & PSNR $\uparrow$ & SSIM $\uparrow$ & LPIPS $\downarrow$ & FID $\downarrow$ & CLIP $\uparrow$ & DISTS $\downarrow$ \\ \hline
        PN & 15.33 & 0.29 & 0.59 & 371.23 & 0.83 & 0.44 & 15.50 & 0.31 & 0.58 & 363.68 & 0.83 & 0.43 \\
        EFT & \textbf{21.28} & 0.69 & 0.34 & 293.36 & 0.87 & 0.33 & \textbf{22.62} & 0.74 & 0.29 & 242.87 & 0.89 & 0.30 \\
        VF & 18.42 & 0.71 & 0.29 & 248.23 & 0.82 & 0.29 & 18.91 & 0.72 & 0.28 & 240.21 & 0.87 & 0.29 \\
        SF & \textbf{21.28} & 0.76 & 0.23 & 187.22 & 0.91 & 0.26 & 22.31 & 0.78 & 0.22 & 175.02 & 0.92 & 0.24 \\
        Ours & 20.95 & \textbf{0.77} & \textbf{0.22} & \textbf{147.65} & \textbf{0.93} & \textbf{0.23} & 22.06 & \textbf{0.79} & \textbf{0.20} & \textbf{134.22} & \textbf{0.94} & \textbf{0.21} \\ \hline
        & \multicolumn{6}{c}{Unseen Instances - 6 views} & \multicolumn{6}{c}{Unseen Categories - 2 views} \\ \hline
        & PSNR $\uparrow$ & SSIM $\uparrow$ & LPIPS $\downarrow$ & FID $\downarrow$ & CLIP $\uparrow$ & DISTS $\downarrow$ & PSNR $\uparrow$ & SSIM $\uparrow$ & LPIPS $\downarrow$ & FID $\downarrow$ & CLIP $\uparrow$ & DISTS $\downarrow$ \\ \hline
        PN & 15.65 & 0.33 & 0.55 & 344.58 & 0.85 & 0.42 & 14.82 & 0.31 & 0.50 & 314.45 & 0.81 & 0.44 \\
        EFT & \textbf{24.47} & 0.80 & 0.23 & 161.78 & 0.93 & 0.25 & \textbf{19.31} & 0.56 & 0.41 & 318.64 & 0.87 & 0.38 \\
        VF & 19.77 & 0.74 & 0.27 & 232.30 & 0.89 & 0.28 & 15.43 & 0.63 & 0.34 & 301.19 & 0.85 & 0.36 \\
        SF & 23.69 & 0.80 & 0.20 & 154.20 & 0.93 & 0.22 & 18.83 & 0.70 & 0.28 & 290.45 & 0.88 & 0.34  \\
        Ours & 23.92 & \textbf{0.82} & \textbf{0.18} & \textbf{116.10} & \textbf{0.95} & \textbf{0.19} & 18.83 & \textbf{0.72} & \textbf{0.23} & \textbf{164.30} & \textbf{0.93} & \textbf{0.26} \\ \hline
    \end{tabular}
    \caption{\textbf{Quantitative comparisons of novel-view synthesis.} We evaluate methods on unseen instances with varying numbers of input images, such as 2, 3, and 6, and on unseen categories with 2 input views. We report the average results across categories for each block.}
    \label{tab:2view_indomain}
\end{table*}

\paragraph{Overall Training.}
We combine all of the losses, including $\mathcal{L}_{\mathrm{C-SDS}},\mathcal{L}_{\text{perp}},\mathcal{L}_{\text{ref}}$, to formulate the objective function of NeRF reconstruction for a specific object.
Once NeRF reconstruction is complete, we can perform volume rendering for novel-view synthesis, and the underlying mesh can be extracted using Marching Cubes~\cite{DBLP:conf/siggraph/LorensenC87}.

\section{Experiment}
In this section, we conduct a qualitative and quantitative evaluation of our approach on the 3D object dataset, CO3Dv2 dataset~\cite{reizenstein21co3d}, to demonstrate its effectiveness. CO3Dv2 dataset is a real-world dataset, which contains 51 common object categories encountered in daily life. We first show the superior quality of novel-view synthesis and 3D reconstruction for unseen object instances in category-specific scenarios with varying numbers of input and then show its out-of-domain generalization ability for unseen categories.

\paragraph{Implementation details.}
For the feature renderer, we follow SparseFusion~\cite{zhou2023sparsefusion} to use three groups of transformer encoders with four 256-dimensional layers to aggregate epipolar features. For the multiview-consistent model, we adopt the Stable Diffusion model v1.5 as our priors.
For NeRF reconstruction, we adapt the threestudio~\cite{threestudio2023}, which is a unified framework for 3D content creation from various inputs, to implement the NeRF reconstruction for specific objects.
We set the weights of the losses with $\lambda_p=100$, $\lambda_c=10$, $\lambda_r=1000$ and $\lambda_m=50$.
NeRF optimization runs for 10,000 steps, which takes about 45 minutes on a single 3090 GPU.

\subsection{Experimental Settings}
\paragraph{Dataset.} We follow the \emph{fewview-train} and \emph{fewview-dev} splits provided by CO3Dv2 dataset~\cite{reizenstein21co3d} for training and evaluation purposes, respectively. For the evaluation of unseen object instances within the same categories, we use the core subset with 10 categories to train the category-specific diffusion model for each category. To assess the out-of-domain generalization ability on unseen categories, we select 10 categories for evaluation and use the remaining 41 categories together for training. Due to the hour-long computation time required for our method, we evaluate only the first 10 object instances of each test split. 

\begin{table}[t]
    \centering
    \small
    \begin{tabular}{c|cc|cc}
    \hline
        & \multicolumn{2}{c|}{Unseen Instances} & \multicolumn{2}{c}{Unseen Categories} \\ \hline
         & CD $\downarrow$ & F-score $\uparrow$ & CD $\downarrow$ & F-score $\uparrow$ \\ \hline
        SF & 0.27 & 0.23 & 0.37 & 0.18 \\
        Ours & \textbf{0.21} & \textbf{0.32} & \textbf{0.27} & \textbf{0.28} \\ \hline
    \end{tabular}
    \caption{\textbf{Quantitative comparison of geometry reconstruction.} Since other baselines only produce images at novel views without 3D representation, we only report the results of ours and SparseFusion. }
    \label{tab:geometry}
\end{table}

\begin{figure}[t]
    \centering
    \includegraphics[scale=0.45]{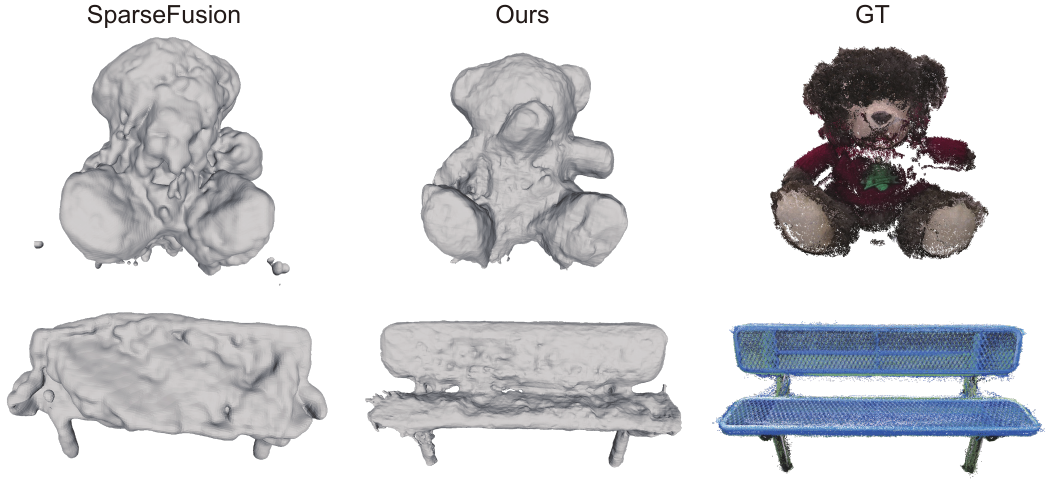}
    \caption{\textbf{Geometry reconstruction using SparseFusion and Ours.} The last column shows the ground-truth point cloud.}
    \label{fig:mesh}
\end{figure}

\paragraph{Baselines.}
We compare our approach with previous state-of-the-art baselines, including PixelNeRF~\cite{DBLP:conf/cvpr/YuYTK21}, ViewFormer~\cite{DBLP:conf/eccv/KulhanekDSB22}, EFT and SparseFusion~\cite{zhou2023sparsefusion}.
PixelNeRF and EFT are regression-based methods that deduce images at novel view by projection feature, where EFT is adapted from GPNR for sparse views settings by ~\cite{zhou2023sparsefusion}.
ViewFormer is a generative model that employs a VQ-VAE codebook and a transformer module for image generation.
Unlike the other methods that directly obtain novel-view synthesis with a single feed-forward pass, SparseFusion is the most relevant baseline to our approach, as it distills the diffusion model prior to NeRF reconstruction.

\begin{figure}[t]
    \centering
    \includegraphics[scale=0.5]{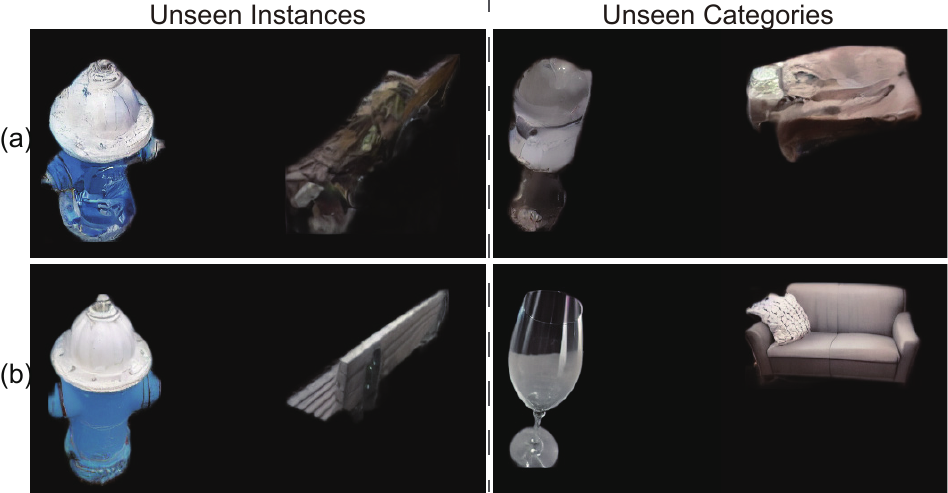}
    \caption{\textbf{Effect of Stable Diffusion priors.} (a) diffusion model from SparseFusion; (b) our diffusion model with Stable Diffusion priors. }
    \label{fig:sd_priors}
\end{figure}

\paragraph{Metrics.}
We adopt several popular image quality assessments (IQA) to evaluate the quality of novel-view synthesis, including PSNR, SSIM, LPIPS~\cite{DBLP:conf/cvpr/ZhangIESW18}, FID~\cite{DBLP:conf/nips/HeuselRUNH17} and DISTS~\cite{DBLP:journals/pami/DingMWS22}. Additionally, since our method can generate plausible results for unobserved regions, the evaluation between them and GT images may not be fair. Thus, we also adopt CLIP embedding similarity~\cite{DBLP:conf/icml/RadfordKHRGASAM21} of generated images with input images. 
Additionally, we evaluate the most commonly used 3D reconstruction quality metrics, including Chamfer Distance and F-score.

\subsection{Qualitative and Quantitative Evaluation}
\paragraph{Unseen Instances: 2 Views.}
We first evaluate our approach with extremely sparse views (i.e. 2 views) for unseen object instances within the same categories. Table~\ref{tab:2view_indomain} demonstrates the quantitative comparison of ours and other baselines, with metrics averaged across 10 categories. We can observe that our method outperforms the others on most image quality metrics except PSNR.
Although ours has a slightly lower PSNR compared to the others, due to its formulation of pixel-wise MSE which favors mean color rendering results (e.g., blurry images), our approach outperforms all of the others in perception metrics (e.g. LPIPS, FID, etc.).
When an image is blurred, the high-frequency details are suppressed, and pixel values become smoother, leading to lower MSE and higher PSNR. 
This phenomenon is also mentioned in SparseFusion~\cite{zhou2023sparsefusion}.
As the qualitative results are shown in Figure~\ref{fig:2view}, benefiting from two proposed key components, our approach achieves both high-quality and more detailed results with 3D consistency.
In addition to novel-view synthesis, we evaluate the quality of geometry reconstruction by extracting underlying mesh from NeRF.
We only compare ours with SparseFusion, while the others (PixelNeRF, EFT, and ViewFormer) lack 3D representation.
From Table~\ref{tab:geometry}, we can find that our approach significantly outperforms SparseFusion by a wide margin, which demonstrates our method's superiority.
Figure~\ref{fig:mesh} also shows the mesh extracted from NeRF, where our results achieve sharper geometry with more details.

\paragraph{Unseen Instances: Varying Views.}
It's obvious that as the number of input views increases, the results of novel-view synthesis and geometry reconstruction improve. 
Table~\ref{tab:2view_indomain} shows the comparison of novel-view synthesis on 3 and 6 input views, which demonstrates that our approach consistently outperforms the others with varying input views. More detailed evaluation results for each category and more qualitative results of novel-view synthesis and explicit geometry can be found in supplementary materials.

\paragraph{Unseen Categories.}
We conduct an experiment to evaluate the generalization ability to unseen categories between ours and the other baselines. 
Table~\ref{tab:2view_indomain} and Table~\ref{tab:geometry} shows the quantitative results of novel-view synthesis and geometry reconstruction. 
When confronted with the unseen categories that are out of the training domain, the performance of the other methods has a significant drop, while ours still maintains good performance, achieving the best results among them.
The priors from Stable Diffusion enable our diffusion model to faithfully generate images of unseen categories.
The last two columns of Figure~\ref{fig:2view} show the novel-view synthesis of these methods. Our approach still can achieve high-quality images with more details, while the others are blurry and somewhat meaningless.
More evaluation of unseen categories (e.g. with varying views) can be found in supplementary materials.

\subsection{Ablation Studies}

\begin{figure}[t]
    \centering
    \includegraphics[scale=0.42]{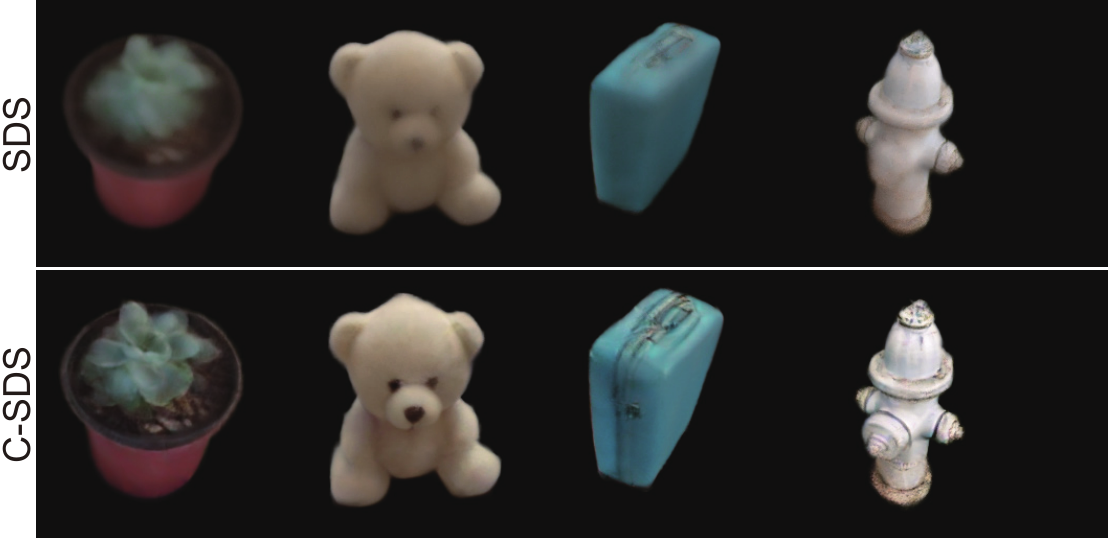}
    \caption{\textbf{Effect of C-SDS to the quality of NVS from NeRF reconstruction.} We can find the results of SDS are blurry and non-detailed in unobserved regions, while ours can generate more details with the same diffusion model.}
    \label{fig:ab_sds}
\end{figure}

\paragraph{Stable Diffusion Priors.}
To evaluate the effect of Stable Diffusion priors, we compare ours and SparseFusion in directly generating novel view images without performing NeRF reconstruction, as shown in Figure~\ref{fig:sd_priors}.
In unseen instances scenario, the diffusion model of SparseFusion can generate images at novel viewpoints consistent with the appearance of input images in a certain way (e.g. the blue hydrant with white head) but fails to achieve high-quality image generation. When the feature map is not reliable in some views, SparseFusion also fails to generate a multiview-consistent image (e.g. the bench). However, our diffusion model can achieve higher-quality of image generation that is more multiview-consistent regarding input images.
In the unseen categories scenario, the diffusion model of SparseFusion fails to generate meaningful images, while our method can be generalized to these objects, benefiting from the Stable Diffusion priors (the last two columns in Figure~\ref{fig:sd_priors}).

\paragraph{C-SDS.}
We also investigate the effect of our distillation strategy on the quality of NeRF reconstruction, by implementing a version of using SDS. 
When using our multiview-consistent diffusion model with SDS, which can provide a more accurate gradient update direction, there is no need for a large CFG, but it's still not enough for detailed results.
In our experiment with setting the CFG value of SDS as 7.5, it can achieve plausible results with successful convergence, but the blur problem is still unsolved, as shown in the first row of Figure~\ref{fig:ab_sds}.
When applying our proposed C-SDS with the same CFG, it's evident that the results show more details, which demonstrates the effectiveness of the method.
The quantitative and more results of ablation studies can be found in supplementary materials.

\subsection{Limitations}
While our method has demonstrated promising results, there are still some limitations to its effectiveness.
The primary failure cases include (1) extremely partial observation of an object in input views; (2) the Janus problem and (3) sometimes difficulty recovering thin structures (e.g., umbrella handles) or self-occlusion parts (e.g., inner of bowl or cup). 
Please refer to the supplementary materials for visual examples.
Furthermore, our approach relies on accurate camera poses, which can be challenging to estimate directly from extremely sparse views, resulting in noisy estimates. 
\section{Conclusion}
In this paper, we introduce {\shorttitle}, a new approach to reconstructing high-quality 3D objects from sparse input views with camera poses. We utilize an epipolar controller to guide a pre-trained diffusion model to generate high-quality images that are 3D consistent with the content of input images, leading to a multiview-consistent diffusion model. Then, we distill the diffusion priors into NeRF optimization in a better way by using a category-score distillation sampling (C-SDS) strategy, resulting in more detailed results.
Experiments demonstrate that our approach can achieve state-of-the-art results with higher quality and more details, even when confronted with open-world, unseen objects.

\section{Acknowledgments}
We sincerely thank the reviewers for their valuable comments. This work was supported by the National Key Research and Development Program of China (No. 2023YFF0905104), the Natural Science Foundation of China (No. 62132012), Beijing Municipal Science and Technology Project (No. Z221100007722001) and Tsinghua-Tencent Joint Laboratory for Internet Innovation Technology.
Shi-Sheng Huang was supported by the Natural Science Foundation of China (Project Number 62202057), State Key Laboratory of Virtual Reality Technology and Systems, Beihang University (No.VRLAB2022B03).

\bibliography{aaai24}

\clearpage
\section{Implementation Details}
\subsection{Feature Renderer}
Since we adapt the Epipolar Feature Transformer (EFT) from SparseFusion~\cite{zhou2023sparsefusion} to feature renderer, we firstly provide a review of EFT and introduce the difference between them.

\paragraph{Review of EFT.}
EFT, which is derived from GPNR, is a feed-forward network that aggregates the features along the epipolar lines of input images and then aggregates the features of aggregated epipolar features from different input views.
Firstly, EFT employs a ResNet18~\cite{DBLP:conf/cvpr/HeZRS16} as an image feature extractor backbone to obtain the features by concatenating intermediate features from the first 4 layers.
Then, for a ray $r$ casting from a query camera viewpoint $\pi$, the EFT uniformly samples $N$ points along the ray direction between the near $d_{near}$ and far $d_{far}$.
The initial features $f_0$ of all sampled points can be concatenated by: (1) projected features tri-linear interpolated from input view images; (2) depths embedding; (3) Pl{\"u}cker coordinates embedding.
Afterward, it employs three transformer modules $T_1,T_2,T_3$ to achieve aggregated features for the final feature map and color image calculation.
The first transformer module is used to combine information from different views. Then an epipolar transformer and a view transformer are used to aggregate features along epipolar lines and different views to achieve a feature map. The process of aggregation is calculated by attention weights $\alpha_k^m$ and $\beta_k$ of transformers and the final feature map $f_c$ can be calculated by:
\begin{equation}
f_c=\sum_{k=1}^{K}\beta_k\left(\sum_{m=1}^{M}\alpha_k^mf_k^m\right)
\end{equation}\label{equ:agg}
where $K$ is the number of input views, $M$ is the number of points sampled along the epipolar lines and $f_k^m$ is the combined features of $k$-th sampled point on $m$-th input image from the output of the first transformer.
Then, a color image $I_f$ can be decoded by an additional linear layer.

\paragraph{Additional Features Embedding.}
In addition to the embedding used in EFT, we incorporate mask embedding and relative camera transformation embedding. Since we focus on object reconstruction, supplementary information indicating which sampled point's projection positions are inside or outside the object is beneficial for the transformer module to pay more attention to the inside features.
Furthermore, the input views of our approach are extremely sparse and there are almost no overlapping areas between two input images, (e.g., the two inputs show the front and the back of an object). In this case, it degrades to a single input situation, and information across different epipolar cannot be effectively combined.
For example, the attention weights $\beta_k$ may have higher values for the other side of the input image, leading to worse view perception.
Thus, a relative camera transformation can let the model learn to pay more attention to the nearer input view.

\begin{figure*}[h]
    \centering
    \includegraphics[scale=0.5]{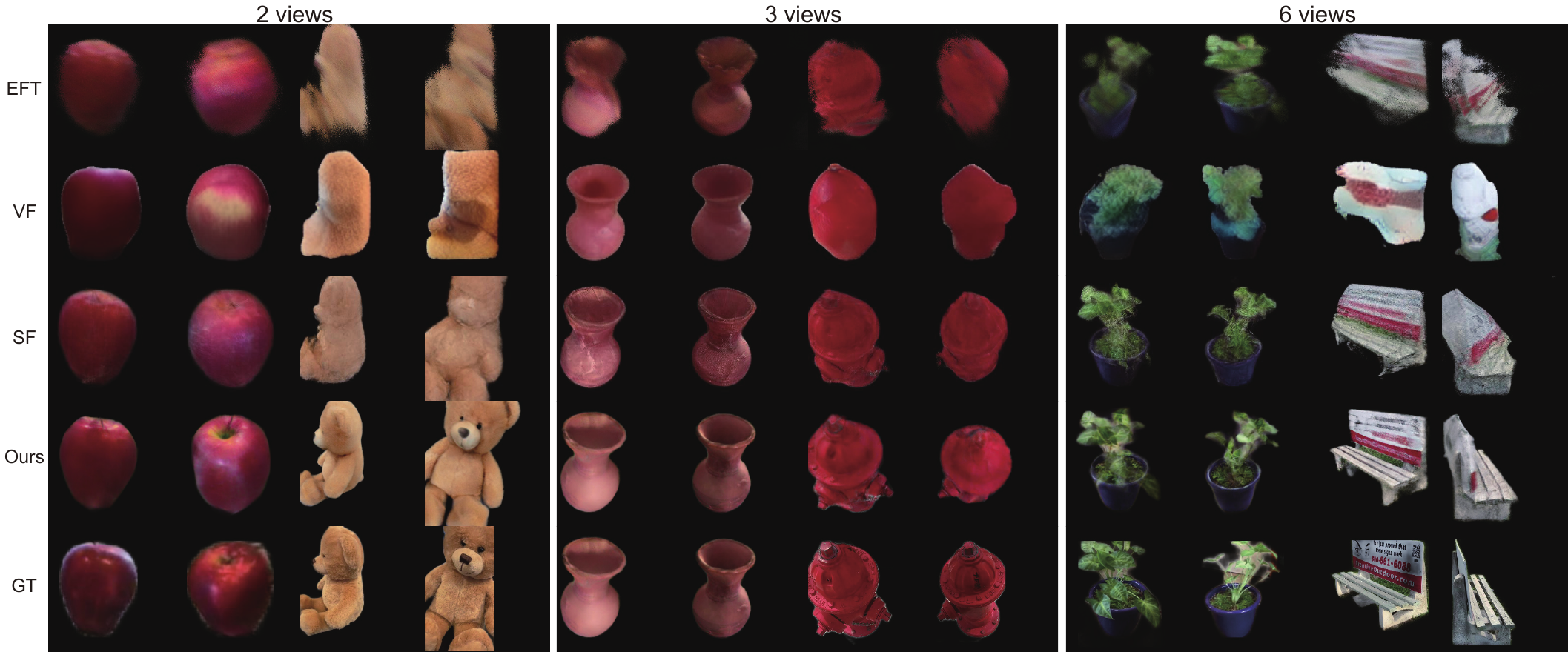}
    \caption{\textbf{Qualitative comparison of novel-view synthesis on unseen instances with a varying number of input views.}}
    \label{fig:instance}
\end{figure*}

\begin{figure*}[h]
    \centering
    \includegraphics[scale=0.5]{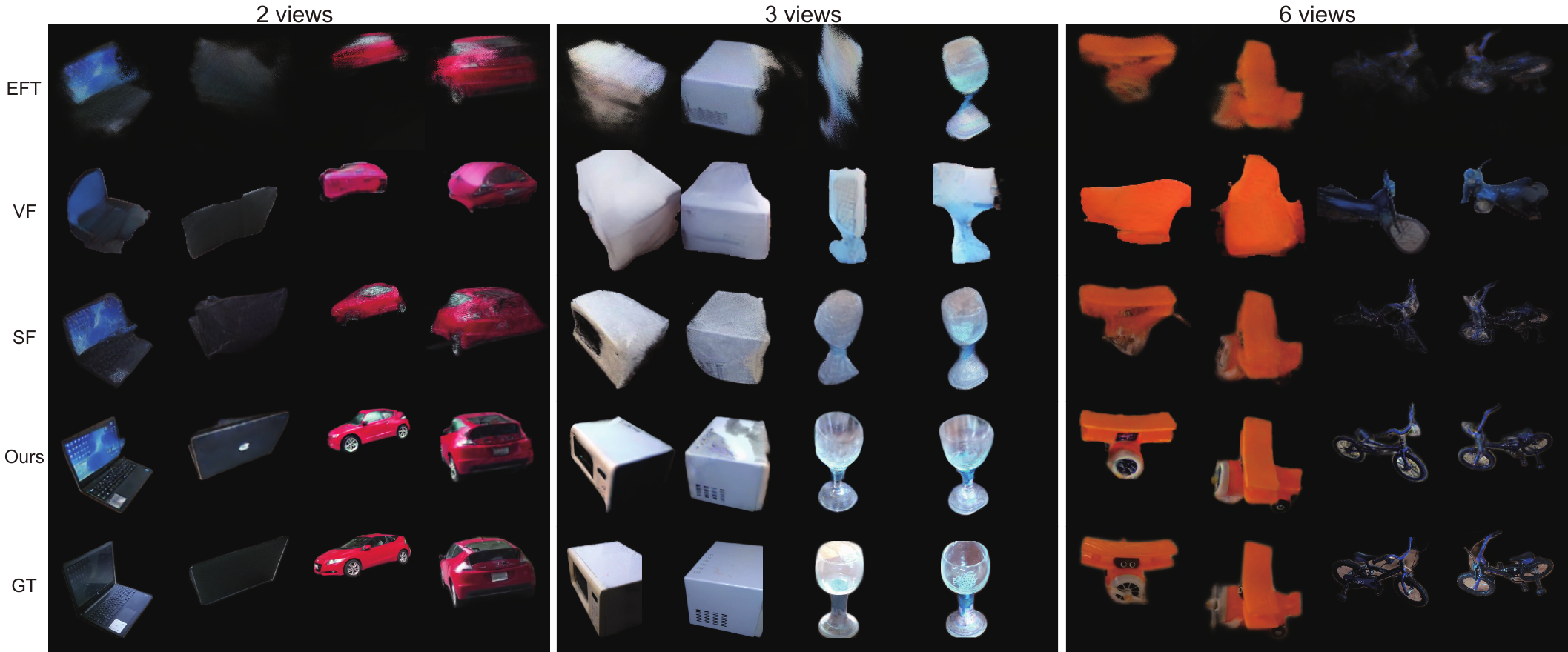}
    \caption{\textbf{Qualitative comparison of novel-view synthesis on unseen categories with a varying number of input views.}}
    \label{fig:category}
\end{figure*}

\paragraph{Aggregated RGB and Depth.}
SparseFusion~\cite{zhou2023sparsefusion} trains EFT by a loss between a decoder color image $I_f$ and ground-truth $I$.
We adopt an aggregated color image $I_{\text{agg}}$ as GPNR does for better generalization ability. To improve the geometry awareness of our feature renderer, we formulate an aggregated depth $D_{\text{agg}}$ with a similar to the aggregated color as:
\begin{equation}
\begin{aligned}
&I_{\text{agg}}=\sum_{k=1}^{K}\beta_k\left(\sum_{m=1}^{M}\alpha_k^m I_k^m\right) \\
    &D_{\text{agg}}=\sum_{k=1}^{K}\beta_k\left(\sum_{m=1}^{M}\alpha_k^m d_k^m\right)
\end{aligned}
\end{equation}
where $I_k^m$ is the RGB value of $k$-th sampled point projected on $m$-th input image, and $d_k^m$ is the sampled depth of this sampled point from the rendering viewpoint. $\alpha_k^m$ and $\beta_k^m$ are the same as Equation~\ref{equ:agg}.
We supervise the $I_{\text{agg}}$ and $D_{\text{agg}}$ by the ground-truth color image and depth.
These can make the transformer modules of feature render to output attention weights with better geometry awareness and generalization ability, which helps for extremely sparse views.

In our implementation, we follow \cite{zhou2023sparsefusion} to sample $N=20$ points along each ray, and set $d_{\text{near}}=s-5$ and $d_{\text{far}}=s+5$ during the training stage, where $s$ is the average distance from scene cameras to origin computed per scene.
Finally, the feature map is rendered at the resolution of $32\times32$ with 256 dimensions for efficiency.

\subsection{Multiview-Consistent Diffusion model}
We implement our diffusion model by using the Diffusers~\cite{von-platen-etal-2022-diffusers}, which is a go-to library for state-of-the-art pre-trained diffusion models. 
We choose Stable Diffusion~\cite{DBLP:conf/cvpr/RombachBLEO22} model as powerful well-studied 2D priors, which includes an encoder $\mathcal{E}(\boldsymbol{x})$, a decoder $\mathcal{D}(\boldsymbol{z})$ and an UNet $\mathcal{U}(\boldsymbol{z})$. The encoder and the decoder are employed to transfer between pixel space $\boldsymbol{x}$ and latent space $\boldsymbol{z}$, and the diffusion process with UNet is employed in latent space.
Specifically, we adopt the network architecture and checkpoint weights from Stable-Diffusion-v1-5~\cite{DBLP:conf/cvpr/RombachBLEO22}.
The epipolar controller is initialized with the same architecture and weights as the encoder blocks and mid-blocks of stable diffusion UNet. 
We employ a convolution layer to align the dimension of the feature map and epipolar controller input.

\begin{figure*}
    \centering
    \includegraphics[scale=0.5]{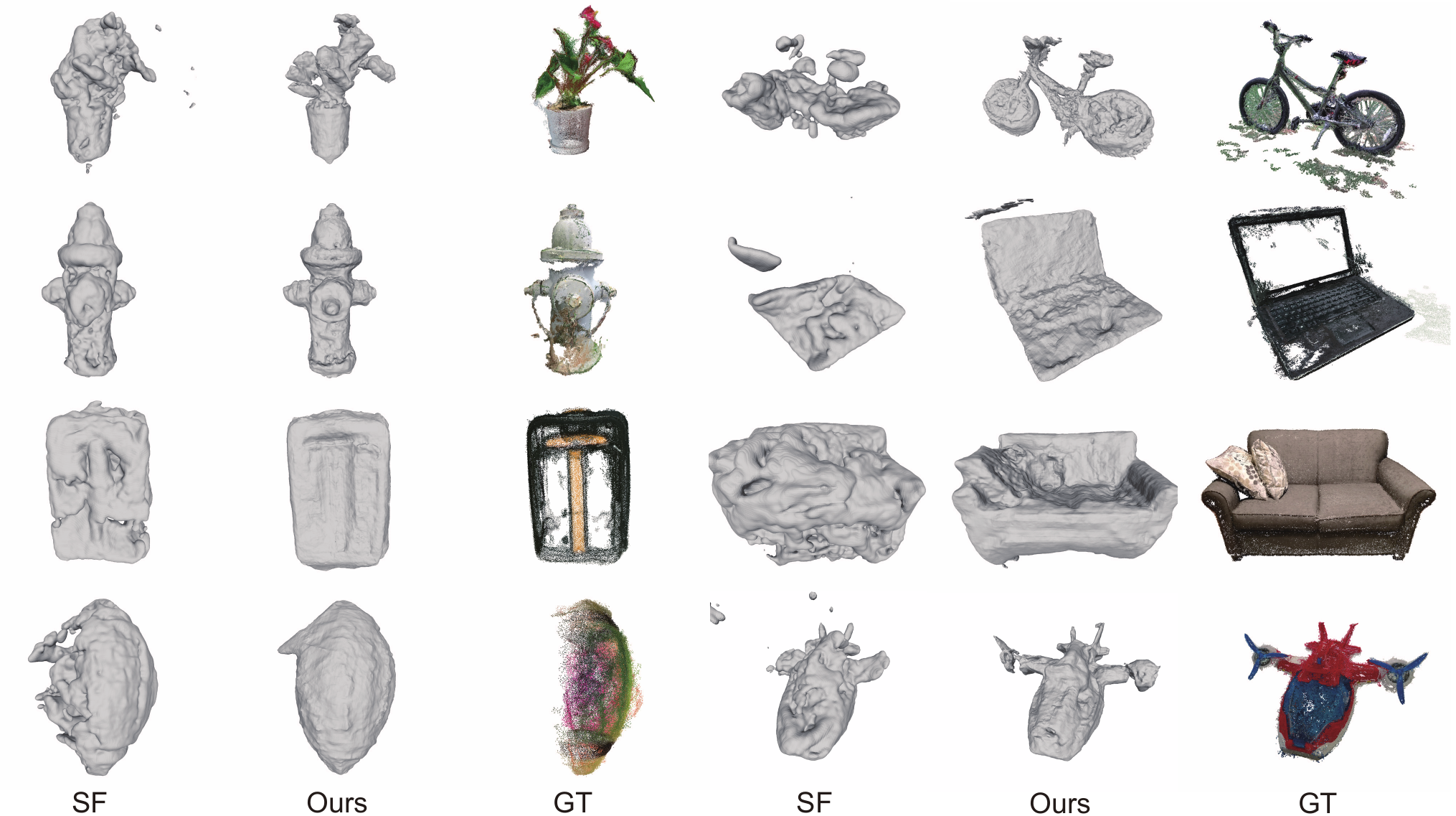}
    \caption{\textbf{Qualitative comparison of geometry reconstruction with SparseFusion.}}
    \label{fig:geometry}
\end{figure*}

\subsection{NeRF Reconstruction}
\paragraph{Rendering from NeRF and Epipolar Features.}
During NeRF reconstruction, we employ two renderers to render images from NeRF representation and render epipolar features from input images, respectively.
When sampling a camera viewpoint at novel viewpoints, we both render an image by the NeRF renderer and an epipolar feature map by the feature renderer. Our multiview-consistent diffusion model takes the rendered feature map as input to guide the NeRF representation through the back-propagation of differentiable volume rendering.
Different from the training stage of the feature renderer, we employ the same near and far values of the feature renderer with the NeRF renderer, which are calculated by the intersections between occupancy grids and ray.

\paragraph{Scene Representation and Rendering.}
For faster rendering and optimization, we utilize the Instance-NGP~\cite{DBLP:journals/tog/MullerESK22} as position encoding, a light-weight MLP with one hidden layer can output density $\sigma$ and color $c$.
We also implement a progressive coarse-to-fine training strategy similar to \cite{li2023neuralangelo}.
During rendering from NeRF representation, we also employ an occupancy grid transmittance estimator during optimization to skip the empty spaces, which can reduce the cost of memory for higher resolution.
The NeRF representation is initialized as a Gaussian sphere for better convergence.
During volume rendering from NeRF, we sample $N=512$ points along the ray implemented by nerfacc~\cite{DBLP:journals/corr/abs-2305-04966} for acceleration.

\paragraph{Additional Regularization.}
In addition to the losses introduced in our paper, we employ three geometry regularization terms on the NeRF reconstruction, which are widely used in other works~\cite{tang2023make,melaskyriazi2023realfusion}, including orientation loss, entropy loss, and sparsity loss.
The orientation loss is proposed by ~\cite{DBLP:conf/cvpr/VerbinHMZBS22} which acts as a penalty on ``foggy'' surfaces, and the other two are following:
\begin{equation}
\begin{aligned}
    &\mathcal{L}_{entropy}=w\cdot log_2(w) - (1 - w)\cdot log_2(1-w) \\
    &\mathcal{L}_{sparsity}=\sqrt{w^2+0.01}
\end{aligned} 
\end{equation}
where $w$ is the cumulative sum of the density.

\section{More Results of Individual Categories}
\subsection{Novel-view Synthesis}
\paragraph{Unseen Instances.}
To evaluate the performance of unseen object instances within the same categories, we experiment on 10 categories of CO3D dataset, including donut, apple, hydrant, vase, cake, ball, bench, suitcase, teddybear and plant. Specifically, we train the models for each category using \emph{fewview-train} split and evaluate them on \emph{fewview-test} split of each category with a varying number of input images (2, 3, and 6), respectively.
We compare our method with PixelNeRF (PN), ViewFormer (VF), EFT, and SparseFusion (SF).
Table~\ref{tab:instance} demonstrates the detailed quantitative results of the novel-view synthesis of each category. Since the metrics of PSNR and SSIM have no perception ability, we pay more attention to the other perception metrics (LPIPS, FID, CLIP, and DISTS), which are more suitable for our evaluation. 
SparseFusion achieves comparable results in most categories with simple geometry or appearance to ours, even though few of the results among them slightly surpass our method (e.g., donut and apple).
However, it has poor performance on those with more complex geometry or appearance (e.g., bench), while our approach can produce higher-quality results on them.
Along with the number of input views increasing, our approach outperforms the others in almost all categories (e.g., except ball with 6 views), consistently with varying numbers of input views.
As Figure~\ref{fig:instance} illustrates some visual results of novel-view synthesis with varying numbers of input views, our approach achieves both high quality and more details than all of the other baselines.

\paragraph{Unseen Categories.}
To evaluate the generalization ability to unseen categories that are out of the training domain, we experiment on 10 categories, including bicycle, car, couch, laptop, microwave, motorcycle, bowl, toyplane, tv, wineglass. Specifically, we train only \emph{one} model on the other 41 categories together, and evaluate them on \emph{fewview-test} split of 10 categories, also with a varying number of input images (2, 3, and 6), respectively.
We can find that when confronted with objects with unseen categories, our approach significantly outperforms the other methods across all test categories. This is mainly due to the prior from Stable Diffusion and generalizable feature renderer. The generalizable feature map can provide a view and appearance perception of the object, and the priors from stable diffusion with the text prompt of the category's name can successfully generate a plausible image regarding the specific input images, resulting in better generalization ability to unseen categories than the others.
Figure~\ref{fig:category} also illustrates some visual results of novel-view synthesis with varying numbers of input views. 
Ours still outperforms the others with higher quality, which demonstrates that our approach can be generalized to unseen categories very well, without further fine-tuning.

\subsection{Geometry Reconstruction}
In addition to the evaluation of novel-view synthesis, we evaluate the performance of geometry reconstruction, since explicit geometry is also essential to many downstream applications. We only compare ours with SparseFusion for geometry reconstruction evaluation, due to the lack of 3D representation from the other methods. Table~\ref{tab:geometry} shows the detailed quantitative results of geometry reconstruction of each category for both unseen instances and unseen categories, with varying numbers of input views.
It demonstrates that our approach outperforms the SparseFusion with a wide margin on both Chamfer Distance and F-score, with all experiment settings.
Figure~\ref{fig:geometry} also shows the qualitative comparison between them, which demonstrates our approach can recover more detailed geometry.
We find that SparseFusion can achieve approximate geometric shapes, where we can recognize them in some way when evaluating unseen instances.
However, in unseen categories scenarios, the results of SparseFusion are much worse, and some of them find it hard to recognize the object from the shape.
In contrast, our approach achieves much better reconstruction results with more details and sharper geometry. And even make up the missing regions in the ground-truth point cloud (the ball in the last row in Figure~\ref{fig:geometry}).
Our approach is based on NeRF representation, which is not suitable for geometry reconstruction.
Thus, some ways to improve the quality of geometry reconstruction may be to utilize the Signed Distance Function (SDF) field by NeuS~\cite{DBLP:conf/nips/WangLLTKW21}, or transfer NeRF representation to tetrahedral SDF grid ~\cite{shen2021dmtet} with coarse-to-fine stage refinement, and we leave it as future work.

\begin{table}[t]
    \centering
    \resizebox{0.5\textwidth}{!}{
    \begin{tabular}{c|cccccc}
        \hline
        & \multicolumn{6}{c}{Unseen Instances} \\ \hline
         & PSNR $\uparrow$ & SSIM $\uparrow$ & LPIPS $\downarrow$ & FID $\downarrow$ & CLIP $\uparrow$ & DISTS $\downarrow$  \\ \hline
         SF & \textbf{20.93} & 0.75 & 0.22 & 145.95 & \textbf{0.93} & \textbf{0.21}  \\
         Ours & 20.51 & \textbf{0.77} & \textbf{0.21} & \textbf{131.99} & \textbf{0.93} & 0.23  \\ \hline
         & \multicolumn{6}{c}{Unseen Categories} \\ \hline
         & PSNR $\uparrow$ & SSIM $\uparrow$ & LPIPS $\downarrow$ & FID $\downarrow$ & CLIP $\uparrow$ & DISTS $\downarrow$  \\ \hline
         SF & 18.17 & 0.68 & 0.29 & 257.19 & 0.90 & 0.27 \\
         Ours & \textbf{18.78} & \textbf{0.72} & \textbf{0.24} & \textbf{138.29} & \textbf{0.93} & \textbf{0.25} \\ \hline
    \end{tabular}
    }
    \caption{\textbf{Quantitative evaluation between the diffusion model of SparseFusion and Ours.}}
    \label{tab:prior}
\end{table}

\begin{table}[t]
    \centering
    \resizebox{0.5\textwidth}{!}{
    \begin{tabular}{c|cccccc|cc}
        \hline
         & PSNR $\uparrow$ & SSIM $\uparrow$ & LPIPS $\downarrow$ & FID $\downarrow$ & CLIP $\uparrow$ & DISTS $\downarrow$ & CD $\downarrow$ & F-score $\uparrow$ \\ \hline
       SDS & \textbf{22.35} & \textbf{0.79} & 0.23 & 175.09 & 0.91 & 0.27 & 0.21 & 0.38  \\
       C-SDS & 22.31 & \textbf{0.79} & \textbf{0.20} & \textbf{132.65} & \textbf{0.94} & \textbf{0.21} & \textbf{0.19} & \textbf{0.39} \\
       \hline
    \end{tabular}
    }
    \caption{\textbf{Quantitative evaluation between SDS and C-SDS on the novel-view synthesis and geometry reconstruction on core-subset of CO3D.}}
    \label{tab:sds}
\end{table}

\begin{table}[h]
    \centering
    \resizebox{0.45\textwidth}{!}{
    \begin{tabular}{c|cccccc}
        \hline
         & PSNR $\uparrow$ & SSIM $\uparrow$ & LPIPS $\downarrow$ & FID $\downarrow$ & CLIP $\uparrow$ & DISTS $\downarrow$ \\ \hline
        SF & 20.29 & 0.74 & 0.25 & 252.38 & 0.89 & 0.30 \\
       SDS & \textbf{21.02} & \textbf{0.77} & 0.22 & 170.77 & 0.92 & 0.26  \\
       C-SDS & 19.85 & 0.76 & \textbf{0.21} & \textbf{147.08} & \textbf{0.93} & \textbf{0.23} \\
       \hline
    \end{tabular}
    }
    \caption{\textbf{Quantitative comparison between SparseFusion, SDS, and our C-SDS on 51 categories.}}
    \label{tab:sds_rebuttal}
\end{table}

\begin{table}[h]
    \centering
    \resizebox{0.45\textwidth}{!}{
    \begin{tabular}{c|cccccc}
        \hline
         & PSNR $\uparrow$ & SSIM $\uparrow$ & LPIPS $\downarrow$ & FID $\downarrow$ & CLIP $\uparrow$ & DISTS $\downarrow$ \\ \hline
        w/o epip & 18.92 & 0.74 & 0.24 & 159.36 & 0.91 & 0.26  \\
        w epip & \textbf{20.51} & \textbf{0.77} & \textbf{0.21} & \textbf{131.99} & \textbf{0.93} & \textbf{0.23} \\
       \hline
    \end{tabular}
    }
    \caption{\textbf{Quantitative evaluation of using or not using epipolar constraint in the controller.}}
    \label{tab:control}
\end{table}

\section{More Analysis for Ablation Study}
\paragraph{Stable Diffusion Priors.}
Table~\ref{tab:prior} shows the quantitative results of novel-view synthesis between the diffusion model of SparseFusion and ours. We report the average of each metric across all categories and varying numbers of inputs. 
In unseen instances experiment, SparseFusion achieves comparable results to ours.
However, in the unseen categories experiment, our approach significantly outperforms it by a large margin. 
The Stable Diffusion priors contain the features of objects from categories that may not have been present in our training domain but are learned from large-scale images. This enables our diffusion model to generalize to unseen categories. 
Moreover, the Stable Diffusion priors also encompass the distribution of high-quality image generation and an additional category prior in the text domain. This further assists our diffusion model in achieving both higher-quality image generation and better multiview consistency.

\begin{figure}[h]
    \centering
    \includegraphics[width=\linewidth]{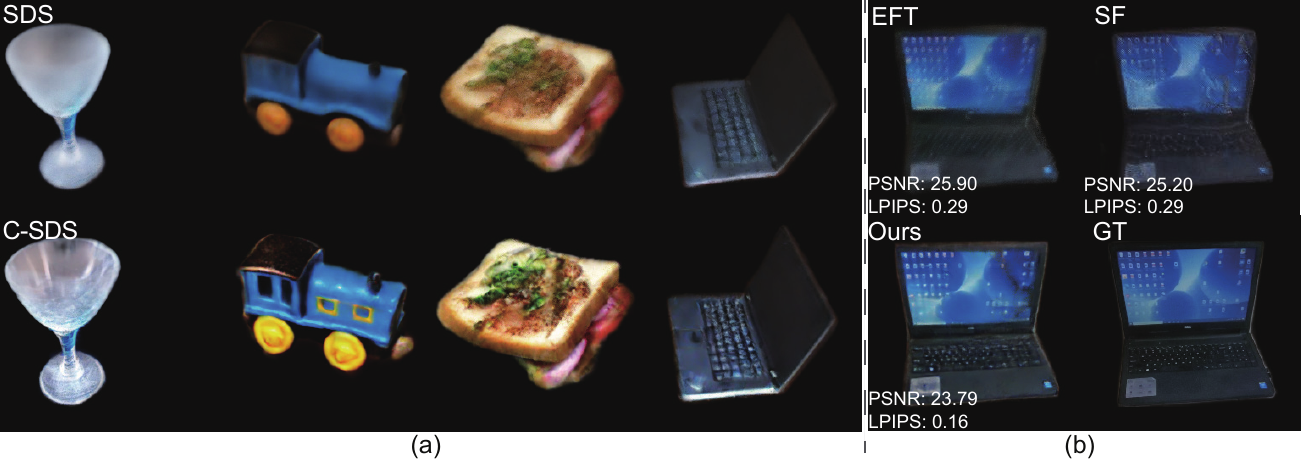}
    \caption{(a) Randomly selected qualitative comparison between SDS and C-SDS. (b) Example of rendering results from EFT, SF, and Ours evaluated on PSNR and LPIPS}
    \label{fig:full}
\end{figure}

\begin{figure}[t]
    \centering
    \includegraphics[scale=0.45]{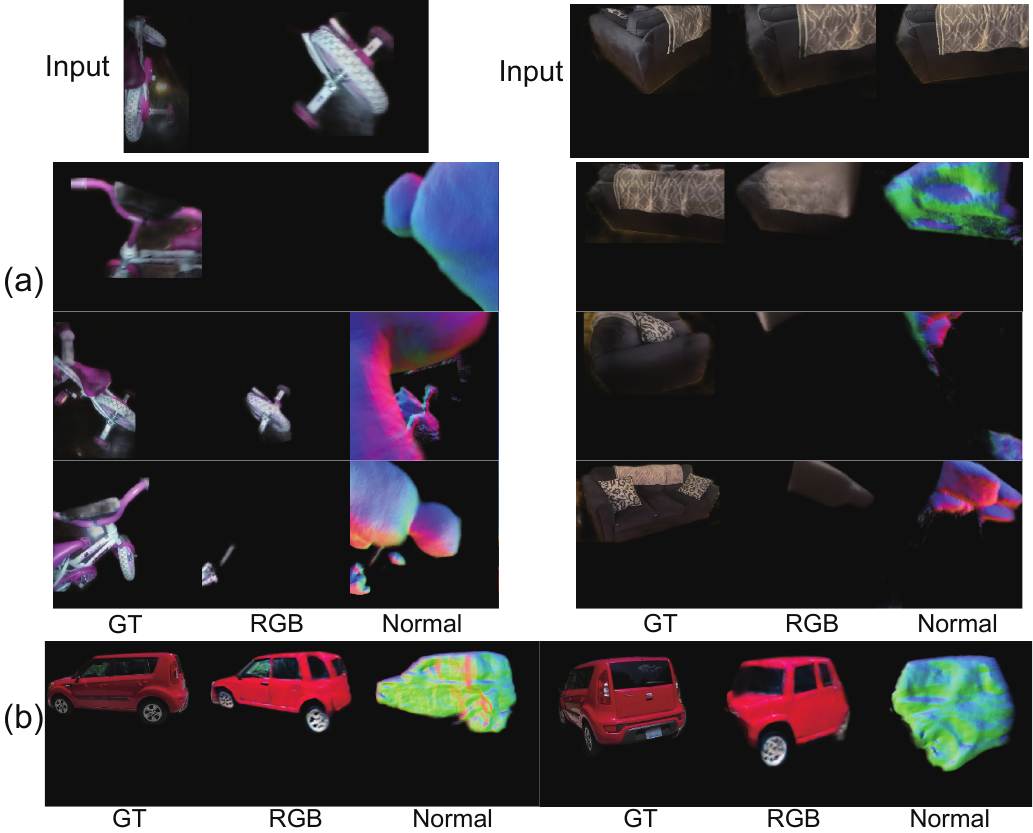}
    \caption{\textbf{Failure cases.} (a) extremely partial observation; (b) Janus problem.}
    \label{fig:failure}
\end{figure}

\paragraph{C-SDS.}
Table~\ref{tab:sds} shows the quantitative results of novel-view synthesis and geometry reconstruction by using SDS and C-SDS, on unseen instances with varying numbers of inputs.
Table~\ref{tab:sds_rebuttal} also demonstrates the comparison between SparseFusion, SDS, and our C-SDS on 51 categories with two input images.
In addition to the problem of blurry images shown in the figure in our paper, The blurring issue of SDS leads to poorer performance on most metrics for novel-view synthesis evaluation, except for PSNR and SSIM.
When utilizing the same diffusion model, our C-SDS can achieve more details from diffusion model priors, resulting in better performance than SDS.
Additionally, we have observed that our C-SDS has minimal impact on the geometry reconstruction quality, with SDS and C-SDS achieving comparable results. 
Furthermore, we have found that the geometry reconstruction results achieved using SDS are significantly superior to those of SparseFusion (with 0.26 Chamfer Distance and 0.24 F-score). This further demonstrates that our multiview-consistent diffusion model has a heightened sense of geometry awareness.
More randomly selected qualitative comparisons between SDS and C-SDS are shown in Figure~\ref{fig:full} (a).

\paragraph{Epipolar Controller Mechanism.}
Given a novel viewpoint, epipolar controller obtains a novel view feature map where each pixel is aggregated by projected features on input images along epipolar lines. The feature map is derived based on 3D structure thus better ensuring consistency. It will go through a U-Net and then be added to the latent maps of Stable Diffusion to control generating the final novel-view image. To show the effectiveness, we remove the epipolar transformer modules and render the feature map using projected features from input views with a volume rendering formulation (like PixelNeRF).
Table~\ref{tab:control} shows that epipolar constraint consistently improves the performance of image generation across all metrics. 

\paragraph{PSNR.} The reason for this is that PSNR favors mean color rendering results (e.g., blurry images) due to its formulation being based on pixel-wise MSE, which is also mentioned in SparseFusion.
When an image is blurred, the high-frequency details are suppressed, and pixel values become smoother. This results in a smaller difference between the original and rendering images, leading to a lower MSE and a higher PSNR.
This phenomenon is also mentioned in SF, and Figure~\ref{fig:full} (b) provides an example from our experiment.
The rendering images of EFT and SF, which are noticeably blurry, achieve higher PSNR values compared to our method, while their performance on LPIPS experiences a significant drop. 

\paragraph{Runtime Efficiency.} Since our method distills diffusion prior into a 3D representation, it requires an additional optimization process compared to feed-forward models, making it challenging to implement for real-time applications and serving as a limitation. Although we cannot generate large-scale 3D objects all at once, our method is still more competitive compared to previous approaches (e.g., SF requires about 65 minutes).
We can utilize multi-GPU to generate them in parallel when faced with this application.
Some potential improvements for computational efficiency could involve using more efficient 3D representation (e.g., 3D Gaussian Splatting) or refining the 3D model based on a feed-forward model prior.

\section{Failure Cases and Limitations}
While our method has demonstrated promising results, there are still some limitations to its effectiveness.
Figure~\ref{fig:failure} shows some failure cases of our approach,
particularly when all input images are too close to the object and only contain a small portion of it, making it difficult for our approach to achieve satisfactory overall results (Figure~\ref{fig:failure} (a)).
Additionally, the Janus problem still occasionally occurs in our results, as depicted in Figure~\ref{fig:failure}(b) where the back of the car shows the appearance of the side.

\section{Supplementary Video}
We provide a video attached with supplementary materials to show the 360-degree visualizations of the other baselines and ours. Please refer to it for more details.

\begin{table*}[t]
    \centering
    \resizebox{\textwidth}{!}{
    \begin{tabular}{c|cc|cc|cc|cc|cc|cc|cc|cc|cc|cc}
        \hline
         & \multicolumn{20}{c}{2 Views}\\ \hline
         & \multicolumn{2}{c|}{Donut} & \multicolumn{2}{c|}{Apple} & \multicolumn{2}{c|}{Hydrant} & \multicolumn{2}{c|}{Vase} & \multicolumn{2}{c|}{Cake} & \multicolumn{2}{c|}{Ball} & \multicolumn{2}{c|}{Bench} & \multicolumn{2}{c|}{Suitcase} & \multicolumn{2}{c|}{Teddybear} & \multicolumn{2}{c}{Plant}  \\ \hline
         & LPIPS $\downarrow$ & FID $\downarrow$ & LPIPS $\downarrow$ & FID $\downarrow$ & LPIPS $\downarrow$ & FID $\downarrow$ & LPIPS $\downarrow$ & FID $\downarrow$ & LPIPS $\downarrow$ & FID $\downarrow$ & LPIPS $\downarrow$ & FID $\downarrow$ & LPIPS $\downarrow$ & FID $\downarrow$ & LPIPS $\downarrow$ & FID $\downarrow$ & LPIPS $\downarrow$ & FID $\downarrow$ & LPIPS $\downarrow$ & FID $\downarrow$\\ \hline
         PN & 0.67 & 374.06 & 0.65 & 306.26 & 0.49 & 355.23 & 0.44 & 314.34 & 0.64 & 420.66 & 0.71 & 360.53 & 0.62 & 390.23 & 0.49 & 382.97 & 0.65 & 417.98 & 0.55 & 390.05 \\
         EFT & 0.32 & 227.03 & 0.34 & 221.76 & 0.24 & 275.57 & 0.27 & 287.28 & 0.40 & 342.09 & 0.37 & 240.34 & 0.41 & 361.97 & 0.31 & 309.97 & 0.35 & 327.01 & 0.38 & 340.54 \\
         VF & 0.29 & 197.35 & 0.26 & 128.33 & 0.23 & 232.44 & 0.22 & 188.77 & 0.33 & 289.76 & 0.32 & 226.12 & 0.31 & 349.23 & 0.27 & 286.24 & 0.33 & 286.78 & 0.31 & 297.25 \\
         SF & \textbf{0.22} & \textbf{114.05} & \textbf{0.20} & 66.15 & 0.16 & 153.11 & \textbf{0.18} & \textbf{140.14} & \textbf{0.28} & 243.81 & \textbf{0.24} & 116.37 & 0.29 & 350.96 & 0.23 & 254.58 & 0.25 & 196.95 & 0.26 & 236.05 \\
         Ours & 0.24 & 123.39 & 0.21 & \textbf{58.99} & \textbf{0.15} & \textbf{126.90} & \textbf{0.18} & 148.01 & 0.30 & \textbf{237.69} & \textbf{0.24} & \textbf{109.29} & \textbf{0.25} & \textbf{205.70} & \textbf{0.19} & \textbf{177.85} & \textbf{0.23} & \textbf{119.87} & \textbf{0.24} & \textbf{168.77} \\ \hline
         & CLIP $\uparrow$ & DISTS $\downarrow$ & CLIP $\uparrow$ & DISTS $\downarrow$ & CLIP $\uparrow$ & DISTS $\downarrow$ & CLIP $\uparrow$ & DISTS $\downarrow$ & CLIP $\uparrow$ & DISTS $\downarrow$ & CLIP $\uparrow$ & DISTS $\downarrow$ & CLIP $\uparrow$ & DISTS $\downarrow$ & CLIP $\uparrow$ & DISTS $\downarrow$ & CLIP $\uparrow$ & DISTS $\downarrow$ & CLIP $\uparrow$ & DISTS $\downarrow$ \\ \hline
         PN & 0.80 & 0.48 & 0.80 & 0.46 & 0.83 & 0.42 & 0.84 & 0.41 & 0.82 & 0.43 & 0.81 & 0.44 & 0.87 & 0.42 & 0.86 & 0.42 & 0.83 & 0.43 & 0.82 & 0.47 \\
         EFT & 0.89 & 0.30 & 0.88 & 0.31 & 0.85 & 0.33 & 0.89 & 0.32 & 0.87 & 0.34 & 0.86 & 0.30 & 0.87 & 0.38 & 0.87 & 0.34 & 0.87 & 0.34 & 0.86 & 0.37 \\
         VF & 0.90 & 0.28 & 0.91 & 0.25 & 0.85 & 0.29 & 0.86 & 0.28 & 0.88 & 0.30 & 0.86 & 0.28 & 0.87 & 0.33 & 0.87 & 0.29 & 0.84 & 0.30 & 0.84 & 0.32 \\
         SF & \textbf{0.93} & \textbf{0.23} & 0.94 & 0.21 & 0.91 & 0.23 & 0.92 & \textbf{0.24} & \textbf{0.91} & 0.27 & \textbf{0.92} & \textbf{0.21} & 0.89 & 0.37 & 0.89 & 0.31 & 0.90 & 0.24 & 0.91 & 0.28 \\
         Ours & \textbf{0.93} & \textbf{0.23} & \textbf{0.96} & \textbf{0.20} & \textbf{0.93} & \textbf{0.19} & \textbf{0.93} & \textbf{0.24} & \textbf{0.91} & \textbf{0.26} & \textbf{0.92} & 0.22 & \textbf{0.92} & \textbf{0.25} & \textbf{0.92} & \textbf{0.22} & \textbf{0.93} & \textbf{0.22} & \textbf{0.93} & \textbf{0.24} \\ \hline
         & \multicolumn{20}{c}{3 Views}\\ \hline
         & \multicolumn{2}{c|}{Donut} & \multicolumn{2}{c|}{Apple} & \multicolumn{2}{c|}{Hydrant} & \multicolumn{2}{c|}{Vase} & \multicolumn{2}{c|}{Cake} & \multicolumn{2}{c|}{Ball} & \multicolumn{2}{c|}{Bench} & \multicolumn{2}{c|}{Suitcase} & \multicolumn{2}{c|}{Teddybear} & \multicolumn{2}{c}{Plant}  \\ \hline
         & LPIPS $\downarrow$ & FID $\downarrow$ & LPIPS $\downarrow$ & FID $\downarrow$ & LPIPS $\downarrow$ & FID $\downarrow$ & LPIPS $\downarrow$ & FID $\downarrow$ & LPIPS $\downarrow$ & FID $\downarrow$ & LPIPS $\downarrow$ & FID $\downarrow$ & LPIPS $\downarrow$ & FID $\downarrow$ & LPIPS $\downarrow$ & FID $\downarrow$ & LPIPS $\downarrow$ & FID $\downarrow$ & LPIPS $\downarrow$ & FID $\downarrow$\\ \hline
         PN & 0.66 & 368.98 & 0.65 & 308.78 & 0.49 & 345.07 & 0.43 & 298.97 & 0.62 & 416.85 & 0.69 & 363.16 & 0.61 & 381.76 & 0.47 & 377.21 & 0.63 & 410.14 & 0.53 & 365.95 \\
         EFT & 0.28 & 181.47 & 0.29 & 163.61 & 0.21 & 230.53 & 0.24 & 238.92 & 0.35 & 287.92 & 0.31 & 180.23 & 0.37 & 331.04 & 0.27 & 270.05 & 0.29 & 248.43 & 0.33 & 296.55 \\
         VF & 0.29 & 196.07 & 0.25 & 117.17 & 0.22 & 224.68 & 0.21 & 187.42 & 0.33 & 280.81 & 0.31 & 216.30 & 0.30 & 344.33 & 0.26 & 280.67 & 0.32 & 269.22 & 0.31 & 285.39 \\
         SF & \textbf{0.22} & 117.39 & \textbf{0.19} & 57.36 & 0.15 & 149.60 & 0.18 & 141.91 & \textbf{0.27} & 233.56 & 0.23 & 98.83 & 0.27 & 332.04 & 0.21 & 224.09 & 0.22 & 167.79 & 0.25 & 227.67 \\
         Ours & \textbf{0.22} & \textbf{110.30} & 0.20 & \textbf{56.31} & \textbf{0.14} & \textbf{117.87} & \textbf{0.17} & \textbf{129.87} & \textbf{0.27} & \textbf{209.09} & \textbf{0.22} & \textbf{93.80} & \textbf{0.22} & \textbf{192.39} & \textbf{0.18} & \textbf{164.42} & \textbf{0.20} & \textbf{106.93} & \textbf{0.23} & \textbf{161.20} \\ \hline
         & CLIP $\uparrow$ & DISTS $\downarrow$ & CLIP $\uparrow$ & DISTS $\downarrow$ & CLIP $\uparrow$ & DISTS $\downarrow$ & CLIP $\uparrow$ & DISTS $\downarrow$ & CLIP $\uparrow$ & DISTS $\downarrow$ & CLIP $\uparrow$ & DISTS $\downarrow$ & CLIP $\uparrow$ & DISTS $\downarrow$ & CLIP $\uparrow$ & DISTS $\downarrow$ & CLIP $\uparrow$ & DISTS $\downarrow$ & CLIP $\uparrow$ & DISTS $\downarrow$ \\ \hline
         PN & 0.81  & 0.47  & 0.81  & 0.45  & 0.83  & 0.41  & 0.85  & 0.40  & 0.83  & 0.43  & 0.82  & 0.43  & 0.87  & 0.40  & 0.86  & 0.42  & 0.84  & 0.42  & 0.83  & 0.46 \\
         EFT & 0.91 & 0.28 & 0.90 & 0.28 & 0.88 & 0.30 & 0.91 & 0.29 & 0.89 & 0.31 & 0.89 & 0.27 & 0.87 & 0.35 & 0.88 & 0.31 & 0.90 & 0.30 & 0.89 & 0.34 \\
         VF & 0.91 & 0.27 & 0.92 & 0.24 & 0.86 & 0.29 & 0.86 & 0.28 & 0.88 & 0.29 & 0.88 & 0.27 & 0.87 & 0.32 & 0.87 & 0.29 & 0.85 & 0.29 & 0.85 & 0.31 \\
         SF & 0.93 & 0.22 & 0.94 & 0.19 & 0.92 & 0.22 & 0.92 & 0.24 & \textbf{0.92} & 0.26 & \textbf{0.94} & \textbf{0.20} & 0.89 & 0.33 & 0.90 & 0.27 & 0.92 & 0.22 & 0.92 & 0.26 \\
         Ours & \textbf{0.94} & \textbf{0.21} & \textbf{0.96} & \textbf{0.18} & \textbf{0.94} & \textbf{0.18} & \textbf{0.93} & \textbf{0.23} & \textbf{0.92} & \textbf{0.24} & \textbf{0.94} & \textbf{0.20} & \textbf{0.92} & \textbf{0.23} & \textbf{0.93} & \textbf{0.21} & \textbf{0.94} & \textbf{0.20} & \textbf{0.94} & \textbf{0.23}  \\ \hline
         & \multicolumn{20}{c}{6 Views}\\ \hline
         & \multicolumn{2}{c|}{Donut} & \multicolumn{2}{c|}{Apple} & \multicolumn{2}{c|}{Hydrant} & \multicolumn{2}{c|}{Vase} & \multicolumn{2}{c|}{Cake} & \multicolumn{2}{c|}{Ball} & \multicolumn{2}{c|}{Bench} & \multicolumn{2}{c|}{Suitcase} & \multicolumn{2}{c|}{Teddybear} & \multicolumn{2}{c}{Plant}  \\ \hline
         & LPIPS $\downarrow$ & FID $\downarrow$ & LPIPS $\downarrow$ & FID $\downarrow$ & LPIPS $\downarrow$ & FID $\downarrow$ & LPIPS $\downarrow$ & FID $\downarrow$ & LPIPS $\downarrow$ & FID $\downarrow$ & LPIPS $\downarrow$ & FID $\downarrow$ & LPIPS $\downarrow$ & FID $\downarrow$ & LPIPS $\downarrow$ & FID $\downarrow$ & LPIPS $\downarrow$ & FID $\downarrow$ & LPIPS $\downarrow$ & FID $\downarrow$\\ \hline
         PN & 0.64 & 355.81 & 0.64 & 291.21 & 0.46 & 312.17 & 0.40 & 274.33 & 0.60 & 397.98 & 0.68 & 350.32 & 0.57 & 375.22 & 0.43 & 359.19 & 0.60 & 407.98 & 0.48 & 321.57 \\
         EFT & 0.22 & 110.54 & 0.21 & 69.22 & 0.15 & 147.47 & 0.20 & 165.32 & 0.28 & 198.95 & 0.24 & 110.57 & 0.30 & 259.92 & 0.21 & 191.85 & 0.22 & 159.42 & 0.26 & 204.50\\
         VF & 0.28 & 194.34 & 0.24 & 106.93 & 0.21 & 212.40 & 0.21 & 189.51 & 0.32 & 273.57 & 0.29 & 207.04 & 0.29 & 338.11 & 0.25 & 274.26 & 0.30 & 251.17 & 0.30 & 275.71 \\
         SF & 0.20 & 113.81 & \textbf{0.17} & 54.37 & 0.14 & 137.18 & 0.17 & 137.46 & 0.26 & 214.34 & 0.21 & \textbf{79.95} & 0.25 & 281.18 & 0.19 & 181.71 & 0.21 & 143.68 & 0.23 & 198.32 \\
         Ours & \textbf{0.19} & \textbf{93.80} & \textbf{0.17} & \textbf{53.41} & \textbf{0.12} & \textbf{106.89} & \textbf{0.16} & \textbf{127.75} & \textbf{0.23} & \textbf{166.95} & \textbf{0.20} & 82.32 & \textbf{0.20} & \textbf{168.25} & \textbf{0.15} & \textbf{133.26} & \textbf{0.18} & \textbf{89.66} & \textbf{0.20} & \textbf{138.67} \\ \hline
         & CLIP $\uparrow$ & DISTS $\downarrow$ & CLIP $\uparrow$ & DISTS $\downarrow$ & CLIP $\uparrow$ & DISTS $\downarrow$ & CLIP $\uparrow$ & DISTS $\downarrow$ & CLIP $\uparrow$ & DISTS $\downarrow$ & CLIP $\uparrow$ & DISTS $\downarrow$ & CLIP $\uparrow$ & DISTS $\downarrow$ & CLIP $\uparrow$ & DISTS $\downarrow$ & CLIP $\uparrow$ & DISTS $\downarrow$ & CLIP $\uparrow$ & DISTS $\downarrow$ \\ \hline
         PN & 0.82 & 0.47 & 0.83 & 0.43 & 0.84 & 0.39 & 0.86 & 0.39 & 0.85 & 0.41 & 0.84 & 0.43 & 0.89 & 0.39 & 0.88 & 0.40 & 0.86 & 0.41 & 0.84 & 0.45 \\
         EFT & 0.94 & 0.23 & 0.94 & 0.22 & 0.92 & 0.24 & 0.93 & 0.26 & 0.93 & 0.26 & 0.93 & 0.22 & 0.90 & 0.32 & 0.92 & 0.26 & 0.94 & 0.24 & 0.92 & 0.28 \\
         VF & 0.91 & 0.27 & 0.93 & 0.23 & 0.88 & 0.28 & 0.88 & 0.28 & 0.89 & 0.29 & 0.90 & 0.26 & 0.88 & 0.31 & 0.89 & 0.28 & 0.87 & 0.28 & 0.86 & 0.30  \\
         SF & 0.94 & 0.21 & 0.95 & 0.17 & 0.93 & 0.21 & 0.93 & 0.23 & 0.93 & 0.25 & \textbf{0.95} & \textbf{0.18} & 0.91 & 0.28 & 0.92 & 0.24 & 0.93 & 0.20 & 0.93 & 0.24 \\
         Ours & \textbf{0.95} & \textbf{0.20} & \textbf{0.97} & \textbf{0.16} & \textbf{0.95} & \textbf{0.17} & \textbf{0.94} & \textbf{0.22} & \textbf{0.94} & \textbf{0.22} & \textbf{0.95} & 0.19 & \textbf{0.94} & \textbf{0.21} & \textbf{0.95} & \textbf{0.18} & \textbf{0.95} & \textbf{0.18} & \textbf{0.95} & \textbf{0.20} \\ \hline
    \end{tabular}
    }
    \caption{\textbf{Quantitative comparison of novel-view synthesis on unseen instances for each category, with varying number of input views (2, 3 and 6).}}
    \label{tab:instance}
\end{table*}

\begin{table*}[t]
    \centering
    \resizebox{\textwidth}{!}{
    \begin{tabular}{c|cc|cc|cc|cc|cc|cc|cc|cc|cc|cc}
    \hline
         & \multicolumn{20}{c}{2 Views}\\ \hline
         & \multicolumn{2}{c|}{Bicycle} & \multicolumn{2}{c|}{Car} & \multicolumn{2}{c|}{Couch} & \multicolumn{2}{c|}{Laptop} & \multicolumn{2}{c|}{Microwave} & \multicolumn{2}{c|}{Motorcycle} & \multicolumn{2}{c|}{Bowl} & \multicolumn{2}{c|}{Toyplane} & \multicolumn{2}{c|}{TV} & \multicolumn{2}{c}{Wineglass}  \\ \hline
         & LPIPS $\downarrow$ & FID $\downarrow$ & LPIPS $\downarrow$ & FID $\downarrow$ & LPIPS $\downarrow$ & FID $\downarrow$ & LPIPS $\downarrow$ & FID $\downarrow$ & LPIPS $\downarrow$ & FID $\downarrow$ & LPIPS $\downarrow$ & FID $\downarrow$ & LPIPS $\downarrow$ & FID $\downarrow$ & LPIPS $\downarrow$ & FID $\downarrow$ & LPIPS $\downarrow$ & FID $\downarrow$ & LPIPS $\downarrow$ & FID $\downarrow$\\ \hline
         PN & 0.56 & 351.03 & 0.52 & 310.50 & 0.51 & 369.59 & 0.49 & 320.08 & 0.56 & 342.70 & 0.50 & 376.82 & 0.56 & 380.98 & 0.50 & 377.76 & 0.53 & 344.72 & 0.50 & 314.45 \\
         EFT & 0.42 & 319.14 & 0.38 & 280.12 & 0.50 & 323.11 & 0.42 & 312.41 & 0.49 & 329.69 & 0.44 & 360.55 & 0.37 & 283.81 & 0.34 & 315.92 & 0.46 & 319.86 & 0.30 & 308.91 \\
         VF & 0.32 & 301.07 & 0.31 & 300.26 & 0.40 & 358.38 & 0.37 & 281.72 & 0.38 & 328.33 & 0.36 & 346.32 & 0.28 & 237.76 & 0.33 & 316.59 & 0.37 & 321.62 & 0.25 & 219.90 \\
         SF & 0.29 & 329.32 & 0.26 & 278.98 & 0.37 & 311.19 & 0.32 & 318.97 & 0.37 & 317.61 & 0.30 & 365.12 & 0.21 & 196.06 & 0.22 & 265.71 & 0.30 & 317.07 & 0.18 & 204.46 \\
         Ours & \textbf{0.23} & \textbf{162.88} & \textbf{0.19} & \textbf{80.67} & \textbf{0.32} & \textbf{270.68} & \textbf{0.23} & \textbf{130.68} & \textbf{0.31} & \textbf{214.97} & \textbf{0.26} & \textbf{206.15} & \textbf{0.16} & \textbf{97.94} & \textbf{0.19} & \textbf{168.73} & \textbf{0.28} & \textbf{236.83} & \textbf{0.16} & \textbf{73.51} \\ \hline
         & CLIP $\uparrow$ & DISTS $\downarrow$ & CLIP $\uparrow$ & DISTS $\downarrow$ & CLIP $\uparrow$ & DISTS $\downarrow$ & CLIP $\uparrow$ & DISTS $\downarrow$ & CLIP $\uparrow$ & DISTS $\downarrow$ & CLIP $\uparrow$ & DISTS $\downarrow$ & CLIP $\uparrow$ & DISTS $\downarrow$ & CLIP $\uparrow$ & DISTS $\downarrow$ & CLIP $\uparrow$ & DISTS $\downarrow$ & CLIP $\uparrow$ & DISTS $\downarrow$ \\ \hline
         PN & 0.86 & 0.47 & 0.85 & 0.45 & 0.90 & 0.42 & 0.84 & 0.45 & 0.87 & 0.40 & 0.86 & 0.50 & 0.83 & 0.43 & 0.79 & 0.51 & 0.89 & 0.41 & 0.81 & 0.44 \\
         EFT & 0.86 & 0.43 & 0.85 & 0.38 & 0.90 & 0.38 & 0.84 & 0.37 & 0.88 & 0.36 & 0.87 & 0.43 & 0.88 & 0.32 & 0.83 & 0.38 & 0.90 & 0.36 & 0.85 & 0.35 \\
         VF & 0.85 & 0.41 & 0.81 & 0.37 & 0.86 & 0.39 & 0.84 & 0.36 & 0.86 & 0.35 & 0.85 & 0.37 & 0.90 & 0.29 & 0.85 & 0.39 & 0.87 & 0.34 & 0.85 & 0.32 \\
         SF & 0.86 & 0.43 & 0.84 & 0.38 & 0.90 & \textbf{0.35} & 0.86 & 0.35 & 0.88 & 0.35 & 0.86 & 0.40 & 0.92 & 0.24 & 0.86 & 0.30 & 0.91 & \textbf{0.30} & 0.88 & 0.28 \\
         Ours & \textbf{0.93} & \textbf{0.30} & \textbf{0.93} & \textbf{0.21} & \textbf{0.92} & 0.37 & \textbf{0.91} & \textbf{0.23} & \textbf{0.92} & \textbf{0.29} & \textbf{0.92} & \textbf{0.28} & \textbf{0.96} & \textbf{0.19} & \textbf{0.92} & \textbf{0.23} & \textbf{0.93} & \textbf{0.30} & \textbf{0.94} & \textbf{0.20} \\ \hline
         & \multicolumn{20}{c}{3 Views}\\ \hline
         & \multicolumn{2}{c|}{Bicycle} & \multicolumn{2}{c|}{Car} & \multicolumn{2}{c|}{Couch} & \multicolumn{2}{c|}{Laptop} & \multicolumn{2}{c|}{Microwave} & \multicolumn{2}{c|}{Motorcycle} & \multicolumn{2}{c|}{Bowl} & \multicolumn{2}{c|}{Toyplane} & \multicolumn{2}{c|}{TV} & \multicolumn{2}{c}{Wineglass}  \\ \hline
         & LPIPS $\downarrow$ & FID $\downarrow$ & LPIPS $\downarrow$ & FID $\downarrow$ & LPIPS $\downarrow$ & FID $\downarrow$ & LPIPS $\downarrow$ & FID $\downarrow$ & LPIPS $\downarrow$ & FID $\downarrow$ & LPIPS $\downarrow$ & FID $\downarrow$ & LPIPS $\downarrow$ & FID $\downarrow$ & LPIPS $\downarrow$ & FID $\downarrow$ & LPIPS $\downarrow$ & FID $\downarrow$ & LPIPS $\downarrow$ & FID $\downarrow$\\ \hline
         PN & 0.46 & 316.75 & 0.51 & 291.66 & 0.54 & 358.22 & 0.52 & 308.25 & 0.53 & 336.86 & 0.52 & 372.39 & 0.54 & 366.09 & 0.49 & 363.66 & 0.52 & 330.17 & 0.49 & 305.69 \\
         EFT & 0.38 & 261.54 & 0.34 & 232.67 & 0.48 & 303.02 & 0.39 & 273.40 & 0.46 & 292.85 & 0.40 & 331.59 & 0.30 & 229.57 & 0.29 & 271.01 & 0.42 & 286.14 & 0.23 & 225.28  \\
         VF & 0.31 & 289.28 & 0.30 & 286.54 & 0.40 & 353.42 & 0.37 & 266.16 & 0.38 & 317.59 & 0.36 & 347.63 & 0.26 & 216.74 & 0.33 & 314.39 & 0.36 & 314.44 & 0.24 & 202.62 \\
         SF & 0.27 & 320.25 & 0.24 & 257.96 & 0.36 & 303.58 & 0.32 & 330.44 & 0.36 & 300.86 & 0.29 & 362.59 & 0.18 & 162.65 & 0.21 & 254.88 & 0.31 & 328.60 & 0.16 & 166.70 \\
         Ours & \textbf{0.22} & \textbf{153.23} & \textbf{0.17} & \textbf{69.97} & \textbf{0.32} & \textbf{265.84} & \textbf{0.22} & \textbf{116.31} & \textbf{0.30} & \textbf{202.69} & \textbf{0.24} & \textbf{198.09} & \textbf{0.14} & \textbf{85.80} & \textbf{0.17} & \textbf{149.40} & \textbf{0.26} & \textbf{225.45} & \textbf{0.14} & \textbf{74.39} \\ \hline
         & CLIP $\uparrow$ & DISTS $\downarrow$ & CLIP $\uparrow$ & DISTS $\downarrow$ & CLIP $\uparrow$ & DISTS $\downarrow$ & CLIP $\uparrow$ & DISTS $\downarrow$ & CLIP $\uparrow$ & DISTS $\downarrow$ & CLIP $\uparrow$ & DISTS $\downarrow$ & CLIP $\uparrow$ & DISTS $\downarrow$ & CLIP $\uparrow$ & DISTS $\downarrow$ & CLIP $\uparrow$ & DISTS $\downarrow$ & CLIP $\uparrow$ & DISTS $\downarrow$ \\ \hline
         PN & 0.87 & 0.48 & 0.85 & 0.44 & 0.92 & 0.38 & 0.86 & 0.42 & 0.82 & 0.41 & 0.87 & 0.47 & 0.84 & 0.42 & 0.80 & 0.51 & 0.90 & 0.40 & 0.82 & 0.44 \\
         EFT & 0.89 & 0.40 & 0.88 & 0.35 & 0.92 & \textbf{0.36} & 0.87 & 0.34 & 0.90 & 0.35 & 0.88 & 0.41 & 0.90 & 0.29 & 0.86 & 0.35 & 0.92 & 0.34 & 0.88 & 0.31 \\ 
         VF & 0.86 & 0.39 & 0.82 & 0.36 & 0.87 & 0.38 & 0.86 & 0.35 & 0.88 & 0.35 & 0.85 & 0.36 & 0.91 & 0.27 & 0.85 & 0.39 & 0.89 & 0.33 & 0.86 & 0.31\\
         SF & 0.87 & 0.39 & 0.86 & 0.35 & 0.91 & \textbf{0.36} & 0.87 & 0.34 & 0.89 & 0.35 & 0.86 & 0.38 & 0.93 & 0.22 & 0.87 & 0.29 & 0.92 & 0.37 & 0.91 & 0.26 \\
         Ours & \textbf{0.94} & \textbf{0.27} & \textbf{0.93} & \textbf{0.19} & \textbf{0.93} & 0.37 & \textbf{0.92} & \textbf{0.22} & \textbf{0.93} & \textbf{0.29} & \textbf{0.93} & \textbf{0.26} & \textbf{0.96} & \textbf{0.17} & \textbf{0.93} & \textbf{0.20} & \textbf{0.94} & \textbf{0.28} & \textbf{0.95} & \textbf{0.19} \\ \hline
         & \multicolumn{20}{c}{6 Views}\\ \hline
         & \multicolumn{2}{c|}{Bicycle} & \multicolumn{2}{c|}{Car} & \multicolumn{2}{c|}{Couch} & \multicolumn{2}{c|}{Laptop} & \multicolumn{2}{c|}{Microwave} & \multicolumn{2}{c|}{Motorcycle} & \multicolumn{2}{c|}{Bowl} & \multicolumn{2}{c|}{Toyplane} & \multicolumn{2}{c|}{TV} & \multicolumn{2}{c}{Wineglass}  \\ \hline
         & LPIPS $\downarrow$ & FID $\downarrow$ & LPIPS $\downarrow$ & FID $\downarrow$ & LPIPS $\downarrow$ & FID $\downarrow$ & LPIPS $\downarrow$ & FID $\downarrow$ & LPIPS $\downarrow$ & FID $\downarrow$ & LPIPS $\downarrow$ & FID $\downarrow$ & LPIPS $\downarrow$ & FID $\downarrow$ & LPIPS $\downarrow$ & FID $\downarrow$ & LPIPS $\downarrow$ & FID $\downarrow$ & LPIPS $\downarrow$ & FID $\downarrow$\\ \hline
         PN & 0.45 & 305.20 & 0.48 & 272.57 & 0.51 & 346.37 & 0.47 & 270.74 & 0.53 & 319.84 & 0.47 & 341.82 & 0.50 & 313.18 & 0.53 & 314.44 & 0.48 & 310.93 & 0.47 & 283.57\\
         EFT & 0.30 & 226.22 & 0.27 & 167.28 & 0.41 & 249.15 & 0.29 & 189.89 & 0.40 & 219.17 & 0.32 & 269.31 & 0.21 & 138.38 & 0.21 & 182.22 & 0.36 & 216.08 & 0.18 & 168.32 \\
         VF & 0.30 & 267.21 & 0.28 & 270.05 & 0.37 & 336.73 & 0.35 & 250.42 & 0.36 & 285.93 & 0.33 & 331.89 & 0.24 & 183.85 & 0.30 & 288.68 & 0.34 & 308.65 & 0.20 & 185.90 \\
         SF & 0.26 & 304.24 & 0.22 & 229.48 & 0.33 & 286.58 & 0.29 & 297.36 & 0.35 & 264.92 & 0.27 & 341.74 & 0.17 & 137.22 & 0.19 & 205.89 & 0.29 & 281.12 & 0.16 & 160.85 \\
         Ours & \textbf{0.19} & \textbf{136.41} & \textbf{0.14} & \textbf{62.36} & \textbf{0.30} & \textbf{217.39} & \textbf{0.18} & \textbf{85.14} & \textbf{0.29} & \textbf{180.00} & \textbf{0.21} & \textbf{156.93} & \textbf{0.11} & \textbf{69.05} & \textbf{0.13} & \textbf{107.69} & \textbf{0.24} & \textbf{177.08} & \textbf{0.13} & \textbf{69.20} \\ \hline
         & CLIP $\uparrow$ & DISTS $\downarrow$ & CLIP $\uparrow$ & DISTS $\downarrow$ & CLIP $\uparrow$ & DISTS $\downarrow$ & CLIP $\uparrow$ & DISTS $\downarrow$ & CLIP $\uparrow$ & DISTS $\downarrow$ & CLIP $\uparrow$ & DISTS $\downarrow$ & CLIP $\uparrow$ & DISTS $\downarrow$ & CLIP $\uparrow$ & DISTS $\downarrow$ & CLIP $\uparrow$ & DISTS $\downarrow$ & CLIP $\uparrow$ & DISTS $\downarrow$ \\ \hline
         PN & 0.88 & 0.44 & 0.82 & 0.42 & 0.93 & 0.36 & 0.89 & 0.41 & 0.89 & 0.38 & 0.89 & 0.46 & 0.86 & 0.40 & 0.82 & 0.48 & 0.93 & 0.40 & 0.83 & 0.43 \\
         EFT & 0.91 & 0.34 & 0.91 & 0.31 & 0.93 & 0.32 & 0.92 & 0.28 & 0.92 & 0.30 & 0.90 & 0.36 & 0.94 & 0.23 & 0.91 & 0.28 & 0.95 & 0.30 & 0.92 & 0.26 \\
         VF & 0.88 & 0.37 & 0.83 & 0.33 & 0.89 & 0.37 & 0.90 & 0.33 & 0.89 & 0.32 & 0.86 & 0.34 & 0.92 & 0.25 & 0.88 & 0.35 & 0.91 & 0.32 & 0.88 & 0.28 \\
         SF & 0.88 & 0.35 & 0.88 & 0.31 & 0.92 & 0.34 & 0.90 & 0.30 & 0.91 & 0.31 & 0.88 & 0.35 & 0.94 & 0.21 & 0.90 & 0.26 & 0.93 & 0.30 & 0.93 & 0.26 \\
         Ours & \textbf{0.95} & \textbf{0.23} & \textbf{0.95} & \textbf{0.18} & \textbf{0.94} & \textbf{0.30} & \textbf{0.94} & \textbf{0.18} & \textbf{0.94} & \textbf{0.26} & \textbf{0.94} & \textbf{0.22} & \textbf{0.97} & \textbf{0.14} & \textbf{0.95} & \textbf{0.17} & \textbf{0.96} & \textbf{0.24} & \textbf{0.95} & \textbf{0.18} \\ \hline
         
    \end{tabular}
    }
    \caption{\textbf{Quantitative comparison of novel-view synthesis on unseen categories for each category, with varying number of input views (2, 3 and 6).}}
    \label{tab:category}
\end{table*}

\begin{table*}[t]
    \centering
    \resizebox{\textwidth}{!}{
    \begin{tabular}{c|cc|cc|cc|cc|cc|cc|cc|cc|cc|cc}
        \hline
        & \multicolumn{20}{c}{Unseen Instances} \\ \hline
         & \multicolumn{20}{c}{2 Views}\\ \hline
         & \multicolumn{2}{c|}{Donut} & \multicolumn{2}{c|}{Apple} & \multicolumn{2}{c|}{Hydrant} & \multicolumn{2}{c|}{Vase} & \multicolumn{2}{c|}{Cake} & \multicolumn{2}{c|}{Ball} & \multicolumn{2}{c|}{Bench} & \multicolumn{2}{c|}{Suitcase} & \multicolumn{2}{c|}{Teddybear} & \multicolumn{2}{c}{Plant}  \\ \hline
         & CD $\downarrow$ & F-score $\uparrow$ & CD $\downarrow$ & F-score $\uparrow$ & CD $\downarrow$ & F-score $\uparrow$ & CD $\downarrow$ & F-score $\uparrow$ & CD $\downarrow$ & F-score $\uparrow$ & CD $\downarrow$ & F-score $\uparrow$ & CD $\downarrow$ & F-score $\uparrow$ & CD $\downarrow$ & F-score $\uparrow$ & CD $\downarrow$ & F-score $\uparrow$ & CD $\downarrow$ & F-score $\uparrow$ \\ \hline
         SF & 0.34 & 0.18 & 0.24 & 0.27 & 0.15 & 0.27 & 0.14 & 0.32 & 0.40 & 0.16 & 0.33 & 0.25 & 0.41 & 0.19 & 0.28 & 0.21 & 0.20 & 0.20 & \textbf{0.21} & 0.20 \\
         Ours & \textbf{0.27} & \textbf{0.26} & \textbf{0.15} & \textbf{0.42} & \textbf{0.13} & \textbf{0.45} & \textbf{0.12} & \textbf{0.42} & \textbf{0.35} & \textbf{0.22} & \textbf{0.27} & \textbf{0.32} & \textbf{0.21} & \textbf{0.27} & \textbf{0.17} & \textbf{0.39} & \textbf{0.15} & \textbf{0.39} & 0.24 & \textbf{0.26} \\ \hline
         & \multicolumn{20}{c}{3 Views}\\ \hline
         & \multicolumn{2}{c|}{Donut} & \multicolumn{2}{c|}{Apple} & \multicolumn{2}{c|}{Hydrant} & \multicolumn{2}{c|}{Vase} & \multicolumn{2}{c|}{Cake} & \multicolumn{2}{c|}{Ball} & \multicolumn{2}{c|}{Bench} & \multicolumn{2}{c|}{Suitcase} & \multicolumn{2}{c|}{Teddybear} & \multicolumn{2}{c}{Plant}  \\ \hline
         & CD $\downarrow$ & F-score $\uparrow$ & CD $\downarrow$ & F-score $\uparrow$ & CD $\downarrow$ & F-score $\uparrow$ & CD $\downarrow$ & F-score $\uparrow$ & CD $\downarrow$ & F-score $\uparrow$ & CD $\downarrow$ & F-score $\uparrow$ & CD $\downarrow$ & F-score $\uparrow$ & CD $\downarrow$ & F-score $\uparrow$ & CD $\downarrow$ & F-score $\uparrow$ & CD $\downarrow$ & F-score $\uparrow$ \\ \hline
         SF & 0.34 & 0.19 & 0.25 & 0.31 & 0.15 & 0.25 & \textbf{0.12} & 0.34 & 0.40 & 0.15 & 0.33 & 0.26 & 0.37 & 0.18 & 0.24 & 0.24 & 0.18 & 0.22 & \textbf{0.21} & 0.18 \\
         Ours & \textbf{0.24} & \textbf{0.31} & \textbf{0.14} & \textbf{0.45} & \textbf{0.11} & \textbf{0.49} & 0.17 & \textbf{0.45} & \textbf{0.28} & \textbf{0.28} & \textbf{0.26} & \textbf{0.36} & \textbf{0.17} & \textbf{0.29} & \textbf{0.16} & \textbf{0.43} & \textbf{0.13} & \textbf{0.47} & 0.22 & \textbf{0.29} \\ \hline
         & \multicolumn{20}{c}{6 Views}\\ \hline
         & \multicolumn{2}{c|}{Donut} & \multicolumn{2}{c|}{Apple} & \multicolumn{2}{c|}{Hydrant} & \multicolumn{2}{c|}{Vase} & \multicolumn{2}{c|}{Cake} & \multicolumn{2}{c|}{Ball} & \multicolumn{2}{c|}{Bench} & \multicolumn{2}{c|}{Suitcase} & \multicolumn{2}{c|}{Teddybear} & \multicolumn{2}{c}{Plant}  \\ \hline
         & CD $\downarrow$ & F-score $\uparrow$ & CD $\downarrow$ & F-score $\uparrow$ & CD $\downarrow$ & F-score $\uparrow$ & CD $\downarrow$ & F-score $\uparrow$ & CD $\downarrow$ & F-score $\uparrow$ & CD $\downarrow$ & F-score $\uparrow$ & CD $\downarrow$ & F-score $\uparrow$ & CD $\downarrow$ & F-score $\uparrow$ & CD $\downarrow$ & F-score $\uparrow$ & CD $\downarrow$ & F-score $\uparrow$ \\ \hline
         SF & 0.32 & 0.22 & 0.26 & 0.30 & 0.13 & 0.26 & 0.14 & 0.36 & 0.38 & 0.18 & 0.33 & 0.27 & 0.26 & 0.20 & 0.23 & 0.28 & 0.19 & 0.23 & 0.21 & 0.18  \\
         Ours & \textbf{0.22} & \textbf{0.38} & \textbf{0.13} & \textbf{0.52} & \textbf{0.09} & \textbf{0.59} & \textbf{0.12} & \textbf{0.48} & \textbf{0.25} & \textbf{0.34} & \textbf{0.25} & \textbf{0.42} & \textbf{0.13} & \textbf{0.39} & \textbf{0.14} & \textbf{0.50} & \textbf{0.11} & \textbf{0.56} & \textbf{0.18} & \textbf{0.36} \\ \hline
         \hline
         & \multicolumn{20}{c}{Unseen Categories} \\ \hline
         & \multicolumn{20}{c}{2 Views}\\ \hline
         & \multicolumn{2}{c|}{Bicycle} & \multicolumn{2}{c|}{Car} & \multicolumn{2}{c|}{Couch} & \multicolumn{2}{c|}{Laptop} & \multicolumn{2}{c|}{Microwave} & \multicolumn{2}{c|}{Motorcycle} & \multicolumn{2}{c|}{Bowl} & \multicolumn{2}{c|}{Toyplane} & \multicolumn{2}{c|}{TV} & \multicolumn{2}{c}{Wineglass}  \\ \hline
         & CD $\downarrow$ & F-score $\uparrow$ & CD $\downarrow$ & F-score $\uparrow$ & CD $\downarrow$ & F-score $\uparrow$ & CD $\downarrow$ & F-score $\uparrow$ & CD $\downarrow$ & F-score $\uparrow$ & CD $\downarrow$ & F-score $\uparrow$ & CD $\downarrow$ & F-score $\uparrow$ & CD $\downarrow$ & F-score $\uparrow$ & CD $\downarrow$ & F-score $\uparrow$ & CD $\downarrow$ & F-score $\uparrow$ \\ \hline
         SF & 0.44 & 0.16 & 0.35 & 0.23 & 0.66 & 0.09 & 0.45 & 0.15 & 0.52 & 0.14 & 0.24 & 0.24 & 0.35 & 0.19 & 0.22 & 0.23 & 0.36 & 0.15 & 0.18 & 0.23 \\
         Ours & \textbf{0.32} & \textbf{0.28} & \textbf{0.24} & \textbf{0.33} & \textbf{0.53} & \textbf{0.16} & \textbf{0.27} & \textbf{0.20} & \textbf{0.28} & \textbf{0.26} & \textbf{0.23} & \textbf{0.27} & \textbf{0.18} & \textbf{0.33} & \textbf{0.19} & \textbf{0.32} & \textbf{0.31} & \textbf{0.27} & \textbf{0.14} & \textbf{0.36} \\ \hline
         & \multicolumn{20}{c}{3 Views}\\ \hline
         & \multicolumn{2}{c|}{Bicycle} & \multicolumn{2}{c|}{Car} & \multicolumn{2}{c|}{Couch} & \multicolumn{2}{c|}{Laptop} & \multicolumn{2}{c|}{Microwave} & \multicolumn{2}{c|}{Motorcycle} & \multicolumn{2}{c|}{Bowl} & \multicolumn{2}{c|}{Toyplane} & \multicolumn{2}{c|}{TV} & \multicolumn{2}{c}{Wineglass}  \\ \hline
         & CD $\downarrow$ & F-score $\uparrow$ & CD $\downarrow$ & F-score $\uparrow$ & CD $\downarrow$ & F-score $\uparrow$ & CD $\downarrow$ & F-score $\uparrow$ & CD $\downarrow$ & F-score $\uparrow$ & CD $\downarrow$ & F-score $\uparrow$ & CD $\downarrow$ & F-score $\uparrow$ & CD $\downarrow$ & F-score $\uparrow$ & CD $\downarrow$ & F-score $\uparrow$ & CD $\downarrow$ & F-score $\uparrow$ \\ \hline
         SF & 0.41 & 0.16 & \textbf{0.25} & 0.28 & \textbf{0.65} & 0.11 & 0.43 & 0.16 & 0.49 & 0.16 & 0.21 & 0.27 & 0.30 & 0.22 & 0.20 & 0.22 & 0.60 & 0.15 & 0.14 & 0.24 \\
         Our & \textbf{0.27} & \textbf{0.33} & 0.42 & \textbf{0.39} & 0.75 & \textbf{0.17} & \textbf{0.20} & \textbf{0.25} & \textbf{0.30} & \textbf{0.25} & \textbf{0.15} & \textbf{0.35} & \textbf{0.15} & \textbf{0.39} & \textbf{0.15} & \textbf{0.38} & \textbf{0.31} & \textbf{0.31} & \textbf{0.11} & \textbf{0.42} \\ \hline
         & \multicolumn{20}{c}{6 Views}\\ \hline
         & \multicolumn{2}{c|}{Bicycle} & \multicolumn{2}{c|}{Car} & \multicolumn{2}{c|}{Couch} & \multicolumn{2}{c|}{Laptop} & \multicolumn{2}{c|}{Microwave} & \multicolumn{2}{c|}{Motorcycle} & \multicolumn{2}{c|}{Bowl} & \multicolumn{2}{c|}{Toyplane} & \multicolumn{2}{c|}{TV} & \multicolumn{2}{c}{Wineglass}  \\ \hline
         & CD $\downarrow$ & F-score $\uparrow$ & CD $\downarrow$ & F-score $\uparrow$ & CD $\downarrow$ & F-score $\uparrow$ & CD $\downarrow$ & F-score $\uparrow$ & CD $\downarrow$ & F-score $\uparrow$ & CD $\downarrow$ & F-score $\uparrow$ & CD $\downarrow$ & F-score $\uparrow$ & CD $\downarrow$ & F-score $\uparrow$ & CD $\downarrow$ & F-score $\uparrow$ & CD $\downarrow$ & F-score $\uparrow$ \\ \hline
         SF & 0.33 & 0.16 & \textbf{0.22} & 0.34 & \textbf{0.61} & 0.14 & 0.31 & 0.23 & 0.43 & 0.19 & 0.15 & 0.30 & 0.23 & 0.26 & 0.17 & 0.28 & 0.43 & 0.17 & 0.15 & 0.26\\
         Our & \textbf{0.19} & \textbf{0.42} & 0.37 & \textbf{0.47} & 0.63 & \textbf{0.24} & \textbf{0.15} & \textbf{0.34} & \textbf{0.29} & \textbf{0.29} & \textbf{0.09} & \textbf{0.44} & \textbf{0.14} & \textbf{0.45} & \textbf{0.12} & \textbf{0.46} & \textbf{0.22} & \textbf{0.38} & \textbf{0.11} & \textbf{0.44} \\ \hline
    \end{tabular}
    }
    \caption{\textbf{Quantitative comparison of geometry reconstruction both on unseen instances and unseen categories for each category, with varying number of input views (2, 3 and 6).}}
    \label{tab:geometry}
\end{table*}

\end{document}